%% file: main.tex
\documentclass[10pt,twocolumn,letterpaper]{article}

\usepackage[pagenumbers]{cvpr} 

\input{preamble}

\definecolor{cvprblue}{rgb}{0.21,0.49,0.74}
\usepackage[pagebackref,breaklinks,colorlinks,allcolors=cvprblue]{hyperref}

\def\confName{CVPR}
\def\confYear{2027}

\title{To Add or Not to Add: Training-Free Contribution Prediction of New Images}

\author{Ümit Mert Çağlar, Alptekin Temizel\\
Graduate School of Informatics\\
METU, Türkiye\\
{\tt\small mecaglar@metu.edu.tr, atemizel@metu.edu.tr}
}

\begin{document}
\maketitle
\input{sec/0_abstract}
\input{sec/1_intro}
\input{sec/2_related}

\input{sec/3_problem}
\input{sec/4_method}
\input{sec/5_benchmark}

\input{sec/6_results}
\input{sec/7_conclusion}
{
    \small
    \bibliographystyle{ieeenat_fullname}
    \bibliography{main}
}

\input{sec/X_suppl}

\end{document}

%% file: preamble.tex
\usepackage{microtype}
\usepackage{amsthm,mathtools,bm}
\usepackage{multirow,array}
\usepackage{colortbl}      
\usepackage{tikz}
\usetikzlibrary{arrows.meta,positioning,fit,calc,backgrounds}



%% file: sec/0_abstract.tex
\begin{abstract}
Vision models rely on the volume and variety of training data, which can be expanded with further data collection and labeling new images or by generating synthetic ones. Adding data is not always beneficial and irrelevant, mislabeled or redundant images can degrade performance, so candidate data should be assessed before use. However, existing data quality metrics are not designed for this setting as they ignore labels, reward redundancy, or require training. 

We introduce the Anticipated Downstream Effect Metric (ADEM), a training-free score that predicts how adding a candidate image set to a trusted training set will change test performance. ADEM scores each candidate image on task relevance, label correctness and uniqueness and then combines these three image-level properties into a single set-level score.

We compare ADEM with 20 existing metrics on 1151 candidate sets from 46 image classification, semantic segmentation and object detection tasks, measuring the gain of each set by training. ADEM ranks the candidate sets with a mean Spearman correlation of 0.85 with measured gain, compared with 0.63 for the strongest baseline, and is the best performing metric on all 46 tasks. Project page is available \href{https://caglarmert.github.io/ADEM.html}{here} and code \href{https://github.com/caglarmert/ADEM}{here}.
\end{abstract}

%% file: sec/1_intro.tex
\section{Introduction}
\label{sec:intro}

Data-driven methods, particularly deep learning, benefit from additional training data but are often constrained by the cost and time required for data collection and annotation. In domains such as medical imaging, remote sensing, and industrial inspection, practitioners therefore augment datasets with data from external sources or synthetic image generators~\cite{caglar2023ulcerative,chen2025scaling,yang2026task,liu2026groundinganomaly,yang2026beyond}. The value of such additions is uncertain before training, as correctly labeled real data is often useful, whereas synthetic data may introduce distribution shifts, label errors, or redundancy that can harm downstream performance.

Training a model on each candidate set directly reveals its utility, but is costly when there are many candidate sets must be evaluated. Also, generative models require metrics to assess their outputs, which is often delegated to fidelity metrics, such as the Fréchet Inception Distance (FID)~\cite{heusel2017gans} and its successors~\cite{binkowski2018demystifying,stein2023exposing,jayasumana2024rethinking}. These metrics quantify the similarity between candidate and reference distributions and are commonly used to assess whether generated or external data resembles the original dataset. However, resemblance does not always imply improved downstream performance.

\begin{figure}[ht]
  \centering
  \includegraphics[width=\columnwidth]{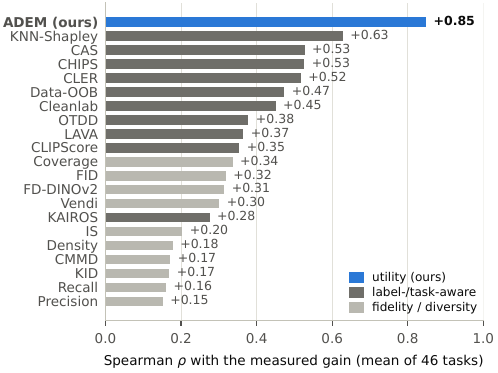}
  \caption{Spearman correlation between each metric and the gain measured by training on~$T\cup D$: mean over 46 tasks.}
  \label{fig:teaser}
\end{figure}

We therefore introduce the Anticipated Downstream Effect Metric (ADEM). Given a small, correctly labeled trusted set and a candidate set, we define the utility of the candidate set as the change in test performance metric, without actually training a downstream model. ADEM is a training-free estimator, calibrated on the trusted set, that estimates the utility of each candidate image with task-relevance, label correctness and uniqueness and then combines these image-level scores into a single set-level score. We benchmark ADEM against 20 existing evaluation metrics across 46 image classification, semantic segmentation and object detection tasks. As shown in \Cref{fig:teaser}, ADEM estimates candidate-set utility more accurately than existing label or task aware performance estimators, as well as fidelity and diversity metrics (detailed in \Cref{tab:main}). It achieves a mean Spearman coefficient~$\rho$ of 0.85, compared with 0.63 for KNN-Shapley and 0.53 for CAS and CHIPS, the strongest baselines. Fidelity and diversity metrics barely track utility ($\rho\le0.34$).

\noindent Our contributions are:
\begin{itemize}
    \item The Anticipated Downstream Effect Metric (ADEM), which predicts whether adding a candidate set or an image to a trusted set will improve a downstream model, without training the downstream model.
    \item A benchmark of 1151 candidate sets on 46 test beds with ground-truth gains obtained by training downstream models. Compared with 20 existing metrics, ADEM ranks the candidate sets with highest correlation with the measured gain on all 46 test beds, including the 32 held-out test beds not used in its design.
\end{itemize}

%% file: sec/2_related.tex
\section{Related Work}
\label{sec:related}

Evaluating candidate data relative to an existing trusted dataset is an important step before using it to improve a downstream task performance. Dataset distances such as the Optimal Transport Dataset Distance (OTDD)~\cite{alvarez2020geometric} and its faster variant sliced-OTDD~\cite{nguyen2025lightspeed} compare labeled datasets without training. LAVA~\cite{just2023lava} evaluates training data without relying on a specific downstream learning algorithm using class-wise Wasserstein distance between training and validation sets as a proxy for validation performance. Data Shapley~\cite{ghorbani2019data} values training samples by their average marginal contribution to downstream performance by multiple retraining runs, estimated by Monte Carlo sampling of subsets. Later variants, such as Data-OOB~\cite{kwon2023dataoob}, reduce the retraining and sampling costs with out-of-bag estimates of a bagging model or compute exact values for k-Nearest-Neighbor models (KNN-Shapley)~\cite{jia2019efficient}, or accumulate values with a single training run~\cite{wang2025data}, or measure each sample's influence on the distance to a clean reference set~\cite{zhu2026kairos}. Influence functions~\cite{koh2017understanding} and datamodels~\cite{ilyas2022datamodels} trace a model’s prediction back to the training points most responsible for it. Recent methods value images during training runs with their marginal contribution to downstream performance~\cite{sun2026online} or with adapting CLIP~\cite{zhuang2026chips}. However, such instance-level valuation methods rank individual samples within a fixed set. Furthermore, Data Shapley can be no better than random unless the utility function has a specific structure with constraints~\cite{wang2024rethinking}. Transferability metrics such as LogME~\cite{you2021logme} and LEEP~\cite{nguyen2020leep} also predict downstream performance without fine-tuning, but they rank pretrained models for a fixed dataset, whereas we rank candidate data for a fixed task.

Numerous metrics also evaluate synthetic or external data. Fréchet Inception Distance (FID)~\cite{heusel2017gans}, Kernel Inception Distance (KID)~\cite{binkowski2018demystifying}, Fréchet Distance with DINOv2 backbone (FD-DINOv2)~\cite{stein2023exposing} and CLIP Maximum Mean Discrepancy (CMMD)~\cite{jayasumana2024rethinking} compare a candidate set to a reference distribution and the Inception Score (IS)~\cite{salimans2016improved} evaluates the candidate set alone. Precision and recall for generative models~\cite{kynkaanniemi2019improved} measure fidelity and diversity as the fraction of samples lying within the $k$-nearest-neighbor manifold of the other set. Density and coverage~\cite{naeem2020reliable} refine these metrics and separate fidelity from diversity and Vendi~\cite{friedman2023vendi} measures diversity. Although widely used to evaluate synthetic data, distributional similarity is the wrong target for training data, as it ignores labels and rewards redundancy. Recent studies confirm that the realism of synthetic images, whether judged by fidelity metrics or improved by newer generators, does not predict their value as training data~\cite{adamkiewicz2026pretty,caglar2026benchmarking}. A training-free metric predicts the utility of generated data for recognition before training~\cite{li2025benchmarking}, but it covers classification only and gives no signed, per-image, label-aware score. Classification Accuracy Score (CAS)~\cite{ravuri2019classification} and studies on training with synthetic images~\cite{he2023synthetic,yan2026makes} move toward downstream utility but they require model training and assess a candidate set in isolation rather than as an addition to existing data.

Related work also spans label-noise detection, data curation and out-of-distribution (OOD) detection. Confident learning~\cite{northcutt2021confident} and margin statistics~\cite{pleiss2020identifying} identify label errors, a recent sample selection method corrects the class and instance level biases of the small-loss criterion~\cite{pan2026debiased}, and a diffusion-based approach recasts label-noise correction as generative label recovery~\cite{hou2025directional}. CLIPScore~\cite{hessel2021clipscore} is widely used as a curation filter~\cite{schuhmann2022laion,gadre2023datacomp}, although different scoring models often give inconsistent quality scores for the same data~\cite{wu2025mixture}. active learning and generative data augmentation improve dataset diversity, while SemDeDup~\cite{abbas2023semdedup} removes semantic duplicates, and perceptual hashing has exposed duplication and train-validation leakage in large geospatial datasets~\cite{adimoolam2026data}. Nearest-neighbor~\cite{sun2022knn} and Mahalanobis~\cite{lee2018simple} distances detect OOD inputs, and conformal inference~\cite{vovk2005algorithmic} converts such scores into calibrated $p$-values. A recent framework detects OOD contamination inside a set~\cite{li2025beyond}. Test error falls roughly as a power law in data size~\cite{hestness2017deep,kaplan2020scaling}, which can be improved with synthetic data for tumor segmentation~\cite{chen2025scaling} while pruning redundant data can even beat it~\cite{sorscher2022beyond}, and fitting such can guide which data to collect, at the cost of training on each domain~\cite{dimlioglu2026scaling}. Coreset and active-learning methods~\cite{sener2018active,coleman2020selection,huseljic2026cleaning} and recent variants~\cite{tan2025data,yu2025mastering} select informative, non-redundant subsets of a pool, and open-set active learning first filters out-of-distribution samples~\cite{schmidt2025joint,zong2026revisiting}. These approaches either target one failure mode, select a subset of a pool under a fixed budget or assume candidate pool is worth sampling. ADEM combines task relevance, label correctness and uniqueness in one score, and because the score is signed, it can not only rank candidate sets but also advise against adding them.

%% file: sec/3_problem.tex
\section{Problem: Utility Relative to a Trusted Set}
\label{sec:problem}

Let $T$ be a small trusted set, correctly labeled and in-distribution with the test set, and $D$ a candidate set of arbitrary origin and size, both with $C$ classes of image labels, segmentation masks or bounding boxes. Each image is assigned either a single label for classification or the majority pixel for segmentation or the majority box for detection.

The utility of $D$ relative to $T$ is the expected change in test performance obtained by adding $D$ to $T$:
\begin{equation}
\Delta(D \mid T) = \mathbb{E}\big[P(T \cup D)\big] - \mathbb{E}\big[P(T)\big],
\label{eq:utility}
\end{equation}
where $P(\cdot)$ is the test performance (macro-F1 for classification, pixel macro-F1 for segmentation, and mAP@0.5 for detection) of a model trained on the given set, and the expectation is over the randomness of training (initialization and data order).

In the benchmark (\cref{sec:benchmark}), expectations are averaged over three seeds. A utility metric shall predict $\Delta$ without training on $D$. It should be signed ($>0$: add, $\le 0$: discard), granular (informative even for a single image), relevance-aware (penalizing task-irrelevant images), label-aware (penalizing wrong labels), uniqueness-aware (discounting exact or near-duplicates of $T$).

%% file: sec/4_method.tex
\section{Anticipating the Downstream Effect of Data}
\label{sec:method}
\begin{figure*}[t]
\centering
\resizebox{\textwidth}{!}{
\begin{tikzpicture}[
  font=\footnotesize,
  >={Stealth[length=2mm]},
  box/.style={draw=black!35, rounded corners=2pt, fill=white, align=center, inner sep=3.5pt, line width=0.5pt},
  databox/.style={box, fill=black!4, text width=2.1cm, minimum height=1.25cm},
  encbox/.style={box, fill=black!4, text width=2.9cm, minimum height=3.6cm},
  facbox/.style={box, text width=4.6cm, align=left, minimum height=1.25cm},
  aggbox/.style={box, minimum height=2.2cm},
  outbox/.style={box, fill=blue!6, draw=blue!60, text width=5.6cm, minimum height=2.2cm},
  arr/.style={->, draw=black!55, line width=0.6pt}
]
\node[encbox] (E) at (0,0) {\textbf{Frozen encoders}\\[5pt]
  {\scriptsize raw + low-pass $\to$ $Z(d)$}\\[4pt]
  {\scriptsize $S(d,t)=\langle Z(d),Z(t)\rangle$}\\[4pt]
  {\scriptsize $k$ nearest in $T$ $\to$ $S(d,T)$}};
\node[databox, left=0.55cm of E, yshift=0.95cm] (T) {\textbf{Trusted set $T$}};
\node[databox, left=0.55cm of E, yshift=-0.95cm] (D) {\textbf{Candidate set $D$}};
\node[facbox, right=0.6cm of E] (Q) {\textcolor{blue!70!black}{\textbf{Label correctness}\hfill$q(d)$}\\[2pt]
  {\scriptsize\begin{tabular}{@{}l@{\;$\to$\;}l@{}}
    probe trained on $T$ & $l(d)$\\
    EM over $D$ & correct labels $\lambda$\\
    \multicolumn{2}{@{}l@{}}{$q(d)$: probability of a correct label}
  \end{tabular}}};
\node[facbox, above=0.22cm of Q] (R) {\textcolor{blue!70!black}{\textbf{Task relevance}\hfill$r(d)$}\\[2pt]
  {\scriptsize\begin{tabular}{@{}l@{$\to$}l@{}}
    rank of $S(d,T)$ among $S(t,T\setminus t)$ & $p(d)$\\
    \multicolumn{2}{@{}l@{}}{$r(d)=\min\{1,\,p(d)/\alpha\}$}
  \end{tabular}}};
\node[facbox, below=0.22cm of Q] (U3) {\textcolor{blue!70!black}{\textbf{Uniqueness}\hfill$u(d)$}\\[2pt]
  {\scriptsize\begin{tabular}{@{}l@{\;$\to$\;}l@{}}
    coverage by $T_c$ & $\kappa_{\max}(d)$\\
    near-copies in $D_c$ & $m(d)$\\
    \multicolumn{2}{@{}l@{}}{$u(d)=\big(1-\kappa_{\max}(d)\big)/m(d)$}
  \end{tabular}}};
\node[aggbox, right=0.6cm of Q] (P) {\textbf{Image utility}\\[4pt]
  $\phi(d)=\underbrace{r(d)\,q(d)\,u(d)}_{\phi_1(d)}$\\[1pt]
  $\phantom{\phi(d)=}-\underbrace{\theta\,r(d)\big(1-q(d)\big)}_{\phi_2(d)}$\\[1pt]
  $\phantom{\phi(d)=}-\underbrace{\mu\big(1-r(d)\big)}_{\phi_3(d)}$\\[4pt]
  {\scriptsize\begin{tabular}{@{}l@{\;}l@{}}
    $\phi_1$: & new, correctly labeled\\
    $\phi_2$: & wrong label\\
    $\phi_3$: & task-irrelevant
  \end{tabular}}};
\node[outbox, right=0.55cm of P] (S) {\textbf{Set score $\mathrm{ADEM}(D,T)$}\\[4pt]
  $\displaystyle\frac1C\sum_{c=1}^{C}\Big[\log\Big(1+\frac{\Phi_{c,1}}{N_c^T}\Big)-\frac{\Phi_{c,2}+\Phi_{c,3}}{N_c^T+N_c^D}\Big]$\\[4pt]
  {\scriptsize\begin{tabular}{@{}l@{\;}l@{}}
    $\Phi_{c,i}$: & $\sum_{d\in D_c}\phi_i(d)$, summed over class $c$\\
    $N_c^T$, $N_c^D$: & class-$c$ images in $T$, $D$\\
    $\mathrm{ADEM}>0$: & add $D$;\quad $\le 0$: do not add
  \end{tabular}}};
\draw[arr] (T.east) -- (T.east -| E.west);
\draw[arr] (D.east) -- (D.east -| E.west);
\draw[arr] (E.east |- R.west) -- (R.west);
\draw[arr] (E.east) -- (Q.west);
\draw[arr] (E.east |- U3.west) -- (U3.west);
\draw[arr] (R.east) -- ++(0.25,0) |- ($(P.west)+(0,0.45)$);
\draw[arr] (Q.east) -- (P.west);
\draw[arr] (U3.east) -- ++(0.25,0) |- ($(P.west)+(0,-0.45)$);
\draw[arr] (P.east) -- (S.west);
\end{tikzpicture}
}
\caption{\textbf{Overview of ADEM.} Every image of the trusted set $T$ (correctly labeled) and the candidate set $D$ (any source, any size down to a single image) is embedded by two frozen encoders, CLIP ViT-B/32 and DINOv2 ViT-S/14 on raw and low-pass views. Each candidate $d$ is then measured against $T$ alone with its \textbf{task relevance} $r(d)$ from a conformal $p$-value, \textbf{label correctness} $q(d)$ from a probe trained on the embeddings of low-pass filtered $T$ and the share of correct labels $\lambda$ estimated by EM over $D$ and its \textbf{uniqueness} $u(d)$ from coverage by $T$ and its near-copies in $D$. The three combine into the image utility $\phi(d)$ and summed per class into the set score.}
\label{fig:pipeline}
\end{figure*}
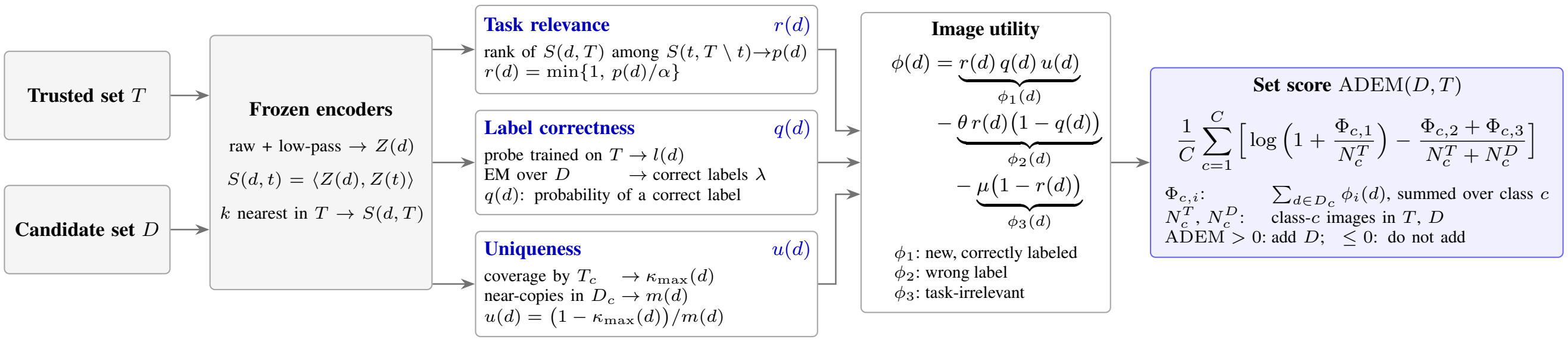

As illustrated in \cref{fig:pipeline}, ADEM starts by embedding every image, trusted or candidate, with CLIP ViT-B/32~\cite{radford2021learning} and DINOv2 ViT-S/14~\cite{oquab2024dinov2}. For a candidate image $d$ (and likewise for a trusted image $t$), $z_{\text{clip,raw}}(d)$ is CLIP's 512-dimensional projected image embedding and $z_{\text{dino,raw}}(d)$ is DINOv2's 384-dimensional CLS token embedding. The image is then blurred by a Gaussian low-pass filter with standard deviation $W/32$, for an image of width $W$, and embedded again, giving $z_{\text{clip,lp}}(d)$ and $z_{\text{dino,lp}}(d)$. Each embedding is normalized to unit length, and the four are concatenated into a 1792-dimensional representation; since the concatenation has norm~2, the factor $\tfrac12$ restores unit length:
\begin{equation}
Z(d)=\tfrac12[z_{\text{clip,raw}}(d);z_{\text{dino,raw}}(d);
       z_{\text{clip,lp}}(d);z_{\text{dino,lp}}(d)]
  \label{eq:latent_representation}
\end{equation}

The low-pass filter removes fine detail, so the blurred views capture an image's coarse content (shapes, layout and color), and noise or compression does not make a genuine candidate look task-irrelevant. The raw views keep the fine detail, so an image whose content was corrupted, for example by heavy noise, cannot pass as a clean candidate behind the blur.

The similarity $S(d,t)$ between a candidate image $d$ and a trusted image $t$, and likewise between any two images, is the cosine similarity of their representations, which equals the mean cosine similarity of the four embeddings, with encoder $e\in\{\text{clip},\text{dino}\}$ and view $v\in\{\text{raw},\text{lp}\}$:

\begin{equation}
  S(d,t)=\langle Z(d),Z(t)\rangle=\frac14\sum_{e,v}\big\langle z_{e,v}(d),z_{e,v}(t)\big\rangle
  \label{eq:sim_single}
\end{equation}

ADEM does not compare $D$ and $T$ as two distributions. Each candidate $d$ is compared with the trusted set on its own, through its mean similarity to the $k$ nearest trusted images, of any class:
\begin{equation}
  S(d,T)=\frac1k\sum_{t\in N_k(d)}S(d,t),
  \label{eq:sim_set}
\end{equation}
where $N_k(d)$ is the set of the $k$ trusted images with the largest similarity $S(d,t)$. Averaging over $k$ neighbors measures how densely the trusted images surround $d$, rather than its similarity to one possibly atypical trusted image. Because $S(d,T)$ is computed per image, ADEM is defined for candidate sets of any size, down to a single image.

\subsection{Task Relevance}
\label{sec:task}

A high similarity $S(d,T)$ suggests that $d\in D$ is relevant to the task. Since the images of some tasks look nearly alike, whereas those of others are diverse, the level of similarity indicative of task relevance is task-dependent.  Rather than using a fixed threshold, ADEM therefore compares $S(d,T)$ with the similarities of the trusted images to one another. For each trusted image $t\in T$, ADEM computes $S(t,T\setminus t)$, its mean similarity to its $k$ most similar images among the other trusted images, as in \eqref{eq:sim_set}. This gives one reference value per trusted image, and these values show how similar genuine images of the task are to $T$. The candidate is then ranked among them:
\begin{equation}
  p(d)=\frac{\big|\{t\in T:\ S(d,T)\ge S(t,T \setminus t)\}\big|+1}{|T|+1}
  \label{eq:pvalue}
\end{equation}
where $|\cdot|$ counts the trusted images that satisfy the condition, and $|T|$ is the number of trusted images. 

Thus, $p(d)$ is the fraction of trusted images that are no more similar to $T$ than $d$ is, known as a conformal $p$-value~\cite{vovk2005algorithmic}. For example, $p(d)=0.5$ means that $d$ is at least as similar to $T$ as half of the trusted images are, and a value near zero means that it is less similar than almost all of them. The 1 added to the numerator and denominator counts $d$ itself and ensures $p(d)>0$. ADEM maps $p(d)$ to a task-relevance weight $r(d) \in (0,1]$ that gives the share of $d$ counted as task-relevant:
\begin{equation}
  r(d)=\min\big\{1,\ p(d)/\alpha\big\},\qquad \alpha=0.05
  \label{eq:ontask}
\end{equation}
A candidate receives full weight $r(d)=1$ when $p(d)\ge\alpha$, that is, when it is at least as similar to $T$ as a fraction $\alpha$ of the trusted images. Above $\alpha$, being more typical earns no extra weight, as an image either belongs to the task or it does not. Below $\alpha$, the weight decreases linearly with $p(d)$. A candidate less similar to $T$ than every trusted image gets the lowest weight, $1/(\alpha(|T|+1))$. The weight is soft, so a borderline image loses little, and $1-r(d)$ is the part of $d$ counted as task-irrelevant in \cref{sec:setscore}. The reference values pool all classes, so task relevance does not use labels.

Because the reference values come from $T$ itself, $\alpha$ has a simple meaning. A genuine candidate, drawn like the trusted images, behaves like one more trusted image: its rank among them is random, so it falls among the least similar fraction $\alpha$ with probability at most $\alpha$. Leaving $d$ out of the reference values keeps this bound, since including $d$ could only raise them and so lower $p(d)$. A genuine image is therefore down-weighted at most $\alpha=5\%$ of the time, and then keeps half of its weight on average, so it loses at most $\alpha/2=2.5\%$ of its weight on average. The argument uses only ranks, so it holds for any task and encoder.

\subsection{Label Correctness}
\label{sec:labels}

ADEM assesses label correctness with a linear probe: a logistic regression trained on the embeddings of low-pass filtered images $z_{\text{clip,lp}}$ and $z_{\text{dino,lp}}$ of the trusted images. Its regularization strength and temperature $\tau$~\cite{guo2017calibration} are selected by cross-validation on $T$ to maximize the likelihood of the held-out trusted labels. Because the probe also learns the class frequencies of $T$, when $T$ is imbalanced, it underestimates the probability of correct labels for rare classes. ADEM therefore removes this class prior~\cite{saerens2002adjusting,menon2021long} by dividing the probe's probabilities by the class frequencies in $T$ and renormalizing. For a candidate $d$ with label $y(d)$, the primary class of $d$ (\cref{sec:problem}), probe logits $f_c(d)$ and $N_c^T$ trusted images of class $c$,
\begin{equation}
  l(d)=\frac{\exp\big(f_{y(d)}(d)/\tau\big)/N_{y(d)}^T}{\sum_{c}\exp\big(f_c(d)/\tau\big)/N_c^T}
  \label{eq:prior}
\end{equation}
is the probability of its label that the probe would give if every class were equally frequent in $T$. On a class-balanced $T$ it is the probe's own probability.

A low $l(d)$ alone does not necessarily indicate a wrong label, since the probe is trained on a small set and may assign low probability to some correct labels. Whether a low probability signals a wrong label depends on the proportion of wrong labels in $D$. ADEM therefore describes the labels of $D$ with a noise model~\cite{natarajan2013learning}, as each label is correct with probability $\lambda$, and is otherwise drawn uniformly at random from the other $C-1$ classes. By Bayes' rule, the probability that the label of $d$ is correct is:
\begin{equation}
  q(d)=\frac{\lambda l(d)}{\lambda l(d)+(1-\lambda)(1-l(d))/(C-1)}
  \label{eq:q}
\end{equation}

The numerator, $\lambda l(d)$, is the probability that the label is correct and the image shows class $y(d)$. The second term of the denominator is the probability that the label is wrong, such as the image shows another class, which the probe believes with probability $1-l(d)$, and $y(d)$ is the wrong class picked, one of $C-1$. For example, in a 10-class task, a label that the probe gives only $l(d)=0.05$ is probably right in a set with $\lambda=0.95$, where $q(d)=0.90$, but probably wrong in a set with $\lambda=0.5$, where $q(d)=0.32$.

The share of correct labels $\lambda$ depends on the source of $D$ and is unknown, so ADEM estimates it from $D$ with expectation-maximization~\cite{saerens2002adjusting}. Starting from $\lambda=0.5$, it computes $q(d)$ for every candidate with the current $\lambda$, and then updates $\lambda$ to a weighted, smoothed average of these values:
\begin{equation}
  \lambda\leftarrow\frac{1+\sum_{d\in D}r(d)\,q(d)}{2+\sum_{d\in D}r(d)}
  \label{eq:em}
\end{equation}
This is repeated until $\lambda$ stops changing. Each candidate is weighted by its task relevance $r(d)$, since a task-irrelevant image states little about the labels of the task. The 1 added to the numerator and the 2 added to the denominator count one imaginary correct and one imaginary wrong label, which keeps a small set from reaching $\lambda=0$ or $1$. A single image has no set to estimate $\lambda$ from, so it uses $\lambda_T$, the value that the same procedure gives for $T$ itself from the probe's held-out probabilities.

\subsection{Uniqueness}
\label{sec:distinct}

Uniqueness quantifies how much a candidate $d$ adds beyond content that $T$ and the other candidates do not already provide. A copy of a trusted image provides no additional information, while multiple copies of a new image should contribute no more than a single copy. ADEM measures image similarity between $x$ and $x'$ using a Gaussian kernel over their representations:
\begin{equation}
   \kappa(x,x')=e^{\frac{S(x,x')-1}{h}}
  \label{eq:kernel}
\end{equation}
which equals 1 for identical images and decays quickly as they differ. We set $h$ to one quarter of the median, over $t \in T$, of the distance $1-S(t,t')$ where $t$ is the nearest same-class trusted neighbor of $t'$ of the same class. Thus, typical same-class neighbors are treated as distinct ($\kappa = e^{-4} \approx 0.02$), while only much closer images, such as near-copies, overlap.

For a candidate image $d$ of class $c$, ADEM measures its coverage by $T$ and the soft count of its near-copies in $D$:
\begin{equation}
   \kappa_{\max}(d)=\max_{t\in T_c} \kappa(d,t),\qquad m(d)=\sum_{d'\in D_c} \kappa(d,d')
  \label{eq:cover}
\end{equation}
where $T_c$ and $D_c$ are the trusted and candidate images labeled $c$. The uniqueness of $d$ is its content not covered by $T$, shared among its near-copies in $D$:
\begin{equation}
  u(d)=\frac{1-\kappa_{\max}(d)}{m(d)}
  \label{eq:distinct}
\end{equation}
Since $d \in D_c$ counts itself, $m(d) \ge 1$ and $u(d) \in [0,1]$. A candidate far from all trusted images and other candidates has $u(d) \approx 1$, an exact copy of a trusted image has $u(d) = 0$, and each of $m$ exact copies of a new image has $u(d) \approx 1/m$, so together they count as one image. Comparisons are restricted to images of the same class, because a copy of a trusted image under a different label is not redundant but mislabeled, which label correctness handles. A single image scored alone has $m(d) = 1$.

\subsection{Image Utility and Set Score}
\label{sec:utility}
\label{sec:setscore}
The three measurements split each candidate into three parts whose weights sum to one: $r(d)q(d)$ for the task-relevant and correctly labeled part, $r(d)(1-q(d))$ for the task-relevant but mislabeled part, and $1-r(d)$ for the task-irrelevant part. Only the first part can improve the model, and only to the extent that it is new, as measured by $u(d)$. ADEM therefore expresses the utility $\phi(d)$ in units of one new, correctly labeled image:
\begin{equation}
  \phi(d)=\underbrace{r(d)\,q(d)\,u(d)}_{\phi_1(d)}
  -\underbrace{\theta\,r(d)\big(1-q(d)\big)}_{\phi_2(d)}
  -\underbrace{\mu\,\big(1-r(d)\big)}_{\phi_3(d)}
  \label{eq:u}
\end{equation}
with $\theta=3$ and $\mu=\tfrac12$, a single image is scored by $\phi(d)$, computed with $\lambda_T$ and $m(d)=1$ and $\phi(d)>0$ advises keeping it.

For a candidate set, ADEM sums each part over the images $D_c$ of class $c$, $\Phi_{c,i}=\sum_{d\in D_c}\phi_i(d)$ for $i=1,2,3$. The set score is
\begin{equation}
  \mathrm{ADEM}(D,T)=\frac1C\sum_{c=1}^{C}\bigg[\log\Big(1+\frac{\Phi_{c,1}}{N_c^T}\Big)
  -\frac{\Phi_{c,2}+\Phi_{c,3}}{N_c^T+N_c^D}\bigg]
  \label{eq:adem}
\end{equation}
where $T$ has $N_c^T$ images of class $c$ and $D$ has $N_c^D$. The first term is the gain of class $c$. Test error typically falls as a power of the number of training images~\cite{hestness2017deep,kaplan2020scaling}, and for a small power, adding $n$ images to $N$ lowers it in proportion to $\log(1+n/N)$. The gain therefore depends on how much a class grows relative to its size, with diminishing returns. The second term measures harm as the weighted share of wrong labels and task-irrelevant images in class $c$ once $D$ is added, as it is the proportion of bad examples, rather than their number, that damages a class~\cite{rolnick2017deep}. Averaging over classes counts every class equally, as macro-F1 and mAP do, so data spread over the classes is worth more than the same amount piled into one.

ADEM has a meaningful zero. In a copy of $T$, every image has $u(d)=0$, so the gain is zero and only a small harm remains. A positive score advises adding $D$, and a zero or negative score advises against it. The set score and the image utility also agree. For a small set, $\log(1+\Phi_{c,1}/N_c^T)\approx \Phi_{c,1}/N_c^T$ and $N_c^T+N_c^D\approx N_c^T$, so the score is about $\frac1C\sum_{d\in D}\phi(d)/N_{y(d)}^T$: each image adds in proportion to its utility, and a single image scores with the sign of $\phi(d)$.

The probe's regularization strength and temperature, its class prior, the reference values of task relevance, the scale $h$ and $\lambda_T$ all depend on the task and computed once from $T$ to be reused for every candidate set. The other settings are the same on every test bed: $k=10$, $\alpha=0.05$, $\theta =3$, $\mu=\tfrac12$ and the quarter in $h$. ADEM trains no network, on $D$ or on any other candidate set. It needs four encoder passes per image, the similarities of the trusted images to one another and of each candidate to $T$ and to the candidates of its class, and one logistic regression fitted on $T$. 

%% file: sec/5_benchmark.tex
\section{Utility Benchmark}
\label{sec:benchmark}
Each of the 46 test beds (\cref{tab:beds}) is a task with a trusted set, a pool of further training images, and a test set. The trusted set is class-balanced except long-tailed datasets. On eight test beds, the generated pools hold synthetic images instead of real ones. We used 14 test beds during ADEM development. The remaining 32, marked with a dagger, were held out. Each test bed has 19 to 42 candidate sets, 1151 in total, consisting of clean images, duplicates, label noise, corruption transforms, task-irrelevant images from other test beds, synthetic images and images from other domains. Clean candidates come in five sizes, from 100 to 2000 images, plus two further independent draws of 500.

\begin{table}[ht]
\centering
\footnotesize
\setlength{\tabcolsep}{2.4pt}
\caption{Spearman correlation $\rho$ with the measured gain; $C$: classes; $|T|$: trusted images; $^\dagger$: held-out test bed.}
\label{tab:beds}
\begin{tabular}{@{}l rrrr c >{\columncolor{blue!8}}c@{}}
\toprule
Bed & $C$ & $|T|$ & Sets & Useful & Baseline & ADEM \\
\midrule
\multicolumn{7}{@{}l}{\textbf{Classification} (macro-F1)}\\
BloodMNIST~\cite{acevedo2020dataset,yang2023medmnist} & 8 & 400 & 23 & 10 & 0.77 & \textbf{0.92} \\
EuroSAT~\cite{helber2019eurosat} & 10 & 500 & 23 & 12 & 0.76 & \textbf{0.87} \\
CIFAR-10~\cite{krizhevsky2009learning,wei2022learning} & 10 & 500 & 41 & 27 & 0.82 & \textbf{0.88} \\
LIMUC~\cite{polat2023limuc} & 4 & 400 & 28 & 16 & 0.66 & \textbf{0.90} \\
RESISC45~\cite{cheng2017remote} & 45 & 450 & 23 & 16 & 0.74 & \textbf{0.93} \\
HAM10000~\cite{tschandl2018ham10000} & 7 & 210 & 24 & 16 & 0.61 & \textbf{0.86} \\
Office-Home~\cite{venkateswara2017deep} & 65 & 325 & 26 & 17 & 0.79 & \textbf{0.89} \\
ARAS400k~\cite{caglar2026grounding} & 6 & 300 & 28 & 14 & 0.71 & \textbf{0.84} \\
LIMUC-128~\cite{polat2023limuc} & 4 & 400 & 28 & 15 & 0.72 & \textbf{0.88} \\
Oxford Pets~\cite{parkhi2012cats}$^\dagger$ & 37 & 370 & 23 & 14 & 0.84 & \textbf{0.90} \\
DTD~\cite{cimpoi2014describing}$^\dagger$ & 47 & 470 & 23 & 15 & 0.79 & \textbf{0.91} \\
GTSRB~\cite{stallkamp2012man}$^\dagger$ & 43 & 430 & 28 & 24 & 0.60 & \textbf{0.67} \\
Galaxy10~\cite{leung2019deep}$^\dagger$ & 10 & 300 & 23 & 17 & 0.54 & \textbf{0.77} \\
Flowers-102~\cite{nilsback2008automated}$^\dagger$ & 102 & 510 & 23 & 18 & 0.70 & \textbf{0.92} \\
FER2013~\cite{goodfellow2013challenges}$^\dagger$ & 7 & 350 & 23 & 19 & 0.68 & \textbf{0.85} \\
KMNIST~\cite{clanuwat2018deep}$^\dagger$ & 10 & 500 & 23 & 14 & 0.68 & \textbf{0.87} \\
CIFAR-10-LT~\cite{krizhevsky2009learning,cao2019learning} & 10 & 499 & 42 & 29 & 0.80 & \textbf{0.92} \\
GTSRB-LT~\cite{stallkamp2012man} & 43 & 431 & 32 & 30 & 0.68 & \textbf{0.89} \\
FER2013-LT~\cite{goodfellow2013challenges} & 7 & 349 & 27 & 18 & 0.65 & \textbf{0.83} \\
HyperKvasir-LT~\cite{borgli2020hyperkvasir} & 23 & 241 & 27 & 23 & 0.58 & \textbf{0.77} \\
\midrule
\multicolumn{7}{@{}l}{\textbf{Segmentation} (pixel macro-F1)}\\
ARAS400k~\cite{caglar2026grounding,caglar2026belde} & 7 & 255 & 31 & 23 & 0.65 & \textbf{0.84} \\
LoveDA~\cite{wang2021loveda}$^\dagger$ & 7 & 301 & 24 & 20 & 0.71 & \textbf{0.87} \\
OpenEarthMap~\cite{xia2023openearthmap}$^\dagger$ & 9 & 310 & 29 & 19 & 0.70 & \textbf{0.95} \\
ISIC 2018~\cite{codella2019skin}$^\dagger$ & 2 & 200 & 22 & 14 & 0.66 & \textbf{0.83} \\
BUSI~\cite{aldhabyani2020dataset}$^\dagger$ & 3 & 120 & 19 & 14 & 0.63 & \textbf{0.82} \\
PanNuke~\cite{gamper2019pannuke}$^\dagger$ & 6 & 223 & 23 & 17 & 0.66 & \textbf{0.88} \\
VOC 2012~\cite{pascal-voc-2012}$^\dagger$ & 21 & 406 & 21 & 8 & 0.73 & \textbf{0.85} \\
FoodSeg103~\cite{wu2021large}$^\dagger$ & 104 & 333 & 23 & 18 & 0.67 & \textbf{0.80} \\
ADE20K~\cite{zhou2017scene}$^\dagger$ & 151 & 244 & 23 & 17 & 0.62 & \textbf{0.69} \\
SUIM~\cite{islam2020semantic}$^\dagger$ & 8 & 217 & 19 & 14 & 0.56 & \textbf{0.74} \\
CamVid~\cite{brostow2009semantic}$^\dagger$ & 12 & 162 & 27 & 18 & 0.84 & \textbf{0.89} \\
\midrule
\multicolumn{7}{@{}l}{\textbf{Detection} (mAP@0.5)}\\
VOC 2007~\cite{pascal-voc-2007}$^\dagger$ & 20 & 391 & 23 & 17 & 0.65 & \textbf{0.89} \\
DIOR~\cite{li2020object,cheng2014multi}$^\dagger$ & 20 & 371 & 23 & 14 & 0.60 & \textbf{0.78} \\
BUSI~\cite{aldhabyani2020dataset}$^\dagger$ & 2 & 100 & 19 & 11 & 0.76 & \textbf{0.82} \\
COCO~\cite{lin2014microsoft}$^\dagger$ & 80 & 358 & 23 & 17 & 0.63 & \textbf{0.86} \\
ExDark~\cite{loh2019getting}$^\dagger$ & 12 & 360 & 23 & 14 & 0.70 & \textbf{0.92} \\
PlantDoc~\cite{singh2020plantdoc}$^\dagger$ & 27 & 270 & 22 & 16 & 0.81 & \textbf{0.94} \\
DocLayNet~\cite{pfitzmann2022doclaynet}$^\dagger$ & 11 & 364 & 23 & 13 & 0.63 & \textbf{0.86} \\
\midrule
\multicolumn{7}{@{}l}{\textbf{Generated pools, classification}}\\
CIFAR-10 EDM~\cite{karras2022elucidating,wang2023better}$^\dagger$ & 10 & 500 & 24 & 15 & 0.83 & \textbf{0.90} \\
CIFAR-10 DDPM~\cite{ho2020denoising,gowal2021improving}$^\dagger$ & 10 & 500 & 24 & 12 & 0.78 & \textbf{0.88} \\
CIFAR-10 SD~\cite{rombach2022high,bird2024cifake}$^\dagger$ & 10 & 500 & 24 & 12 & 0.80 & \textbf{0.89} \\
GTSRB Synset~\cite{sielemann2024synset}$^\dagger$ & 43 & 430 & 24 & 21 & 0.50 & \textbf{0.58} \\
ARAS400k CNet~\cite{zhang2023adding}$^\dagger$ & 6 & 300 & 24 & 4 & 0.78 & \textbf{0.88} \\
\midrule
\multicolumn{7}{@{}l}{\textbf{Generated pools, segmentation}}\\
ARAS400k CNet~\cite{zhang2023adding}$^\dagger$ & 7 & 255 & 27 & 19 & 0.70 & \textbf{0.91} \\
CamVid SYNTHIA~\cite{ros2016synthia}$^\dagger$ & 12 & 162 & 24 & 12 & 0.78 & \textbf{0.91} \\
OpenEarthMap SynRS3D~\cite{song2024synrs3d}$^\dagger$ & 9 & 310 & 25 & 3 & 0.50 & \textbf{0.58} \\
\midrule
Total / Mean & & & 1151 & 746 & 0.70 & 0.85 \\
\bottomrule
\end{tabular}
\end{table}

We train the learner on $T$ and on each $T\cup D$ with three seeds each under a fixed budget of images seen, and estimate $\Delta(D\mid T)$ as the difference of their mean test performance. For classification, we use a 0.39M-parameter CNN trained from scratch at $32\times32$ pixels (500 steps, batch size 128) for BloodMNIST, EuroSAT, CIFAR-10, LIMUC, KMNIST and CIFAR-10-LT and an ImageNet-pretrained ResNet-18~\cite{he2016deep} fine-tuned at $96\times96$ pixels (300 steps, batch size 64) for the remaining 16 datasets. For segmentation, we use an FPN~\cite{lin2017feature} with an ImageNet-pretrained ResNet-18 backbone at $128\times128$ pixels (1200 steps, batch size 16). For detection, we use Faster R-CNN~\cite{ren2015faster} with an ImageNet-pretrained MobileNetV3~\cite{howard2019searching} backbone at $256\times256$ pixels (2400 steps, batch size 8). We consider a candidate useful when its gain exceeds twice its standard error across random seeds. We compare ADEM with 20 metrics (\cref{tab:main}), each oriented such that higher values indicate better data.

We report nine evaluations, each averaged over the test beds. The first four measure how well a metric ranks the candidates of a test bed. The Spearman correlations $\rho_{\text{all}}$ and $\rho_{\text{fix}}$ with $\Delta$ are computed over all candidates and over the fixed-size candidates, respectively. The fixed-size candidates exclude the size sweeps, so that the number of images alone cannot determine the ranking. AUROC measures how well the metric separates useful candidates, whose gain exceeds two standard errors, from the rest. Directions measures how often the metric moves relative to the clean draws in the same direction as the gain, as up, down, or within noise. The next two evaluations assess performance with limited data. $\rho_{@10}$ is $\rho_{\text{fix}}$ when the metric receives only 10 randomly sampled images per candidate, averaged over 20 draws. $\text{AUROC}_1$ scores each image on its own and measures how well it separates images of a clean set that helped from images of corrupted sets that did not. 

The last three evaluations concern decisions. Top-1 is the gain of the fixed-size candidate ranked first, as a percentage of the best gain among them. Pool cleaning, on the 21 classification test beds with a balanced $T$, trains on the images that a per-image score keeps at its own threshold from a corrupted pool and reports the gain over training on $T$ alone. Finally, we report the number of test beds on which each metric has the highest $\rho_{\text{all}}$.

%% file: sec/6_results.tex
\section{Results}
\label{sec:results}

\begin{table*}[t]
\centering
\small
\setlength{\tabcolsep}{4.2pt}
\caption{Comparison with existing metrics, averaged over the 46 test beds and 1151 candidate sets. Dir.: directions; Pool $\Delta$F1: macro-F1 points gained by pool cleaning; \#Best: test beds where the metric has the highest $\rho_{\text{all}}$; $^*$: scored on pairs of images; \textbf{Best} and \underline{second best}.}
\label{tab:main}
\begin{tabular}{@{}l l cccc cc ccc@{}}
\toprule
 & & \multicolumn{4}{c}{\emph{Ranking candidate sets}} & \multicolumn{2}{c}{\emph{Few images}} & \multicolumn{3}{c}{\emph{Decisions}} \\
\cmidrule(lr){3-6}\cmidrule(lr){7-8}\cmidrule(l){9-11}
Metric & Type & $\rho_{\text{all}}\uparrow$ & $\rho_{\text{fix}}\uparrow$ & AUROC$\uparrow$ & Dir.\,\%$\uparrow$ & $\rho_{@10}\uparrow$ & AUROC$_1\uparrow$ & Top-1\,\%$\uparrow$ & Pool $\Delta$F1$\uparrow$ & \#Best$\uparrow$ \\
\midrule
FID~\cite{heusel2017gans} & fidelity & 0.32 & 0.20 & 0.59 & 54 & 0.17 & 0.68$^*$ & 0 & -- & 0 \\
KID~\cite{binkowski2018demystifying} & fidelity & 0.17 & 0.13 & 0.56 & 53 & 0.16 & 0.75$^*$ & 0 & -- & 0 \\
FD-DINOv2~\cite{stein2023exposing} & fidelity & 0.31 & 0.19 & 0.59 & 52 & 0.19 & 0.66$^*$ & 0 & -- & 0 \\
CMMD~\cite{jayasumana2024rethinking} & fidelity & 0.17 & 0.11 & 0.55 & 52 & 0.14 & 0.76$^*$ & 0 & -- & 0 \\
IS~\cite{salimans2016improved} & fidelity & 0.20 & 0.13 & 0.57 & 45 & 0.09 & 0.59$^*$ & -2 & -- & 0 \\
Precision~\cite{kynkaanniemi2019improved} & fidelity & 0.15 & 0.10 & 0.53 & 48 & 0.10 & 0.63 & 0 & -3.3 & 0 \\
Density~\cite{naeem2020reliable} & fidelity & 0.18 & 0.12 & 0.54 & 50 & 0.14 & 0.67 & 11 & -3.5 & 0 \\
\midrule
Recall~\cite{kynkaanniemi2019improved} & diversity & 0.16 & 0.15 & 0.57 & 47 & 0.16 & 0.70$^*$ & 0 & -- & 0 \\
Coverage~\cite{naeem2020reliable} & diversity & 0.34 & 0.20 & 0.59 & 55 & 0.15 & 0.67 & 9 & -3.5 & 0 \\
Vendi~\cite{friedman2023vendi} & diversity & 0.30 & 0.18 & 0.61 & 48 & 0.13 & 0.60$^*$ & -6 & -- & 0 \\
\midrule
OTDD~\cite{alvarez2020geometric} & dataset dist. & 0.38 & 0.29 & 0.63 & 61 & 0.23 & 0.76$^*$ & 0 & -- & 0 \\
LAVA~\cite{just2023lava} & data valuation & 0.37 & 0.26 & 0.61 & 60 & 0.22 & 0.71 & 0 & -2.5 & 0 \\
CLIPScore~\cite{hessel2021clipscore} & curation filter & 0.35 & 0.37 & 0.69 & 58 & 0.32 & 0.71 & \underline{44} & -0.1 & 0 \\
CHIPS~\cite{zhuang2026chips} & data selection & 0.53 & 0.47 & 0.69 & 61 & 0.04 & 0.57 & 33 & -0.7 & 0 \\
CAS~\cite{ravuri2019classification} & train $D$, test $T$ & 0.53 & 0.48 & 0.71 & 64 & 0.06 & 0.70$^*$ & 6 & -- & 0 \\
CLER~\cite{li2025benchmarking} & centres $D$, test $T$ & 0.52 & 0.48 & 0.72 & 65 & 0.40 & 0.72$^*$ & 13 & -- & 0 \\
Cleanlab~\cite{northcutt2021confident} & label quality & 0.45 & 0.48 & 0.71 & 65 & 0.42 & 0.77 & 1 & +0.4 & 0 \\
Data-OOB~\cite{kwon2023dataoob} & data valuation & 0.47 & 0.44 & 0.66 & 63 & 0.31 & 0.44 & 7 & \underline{+0.5} & 0 \\
KAIROS~\cite{zhu2026kairos} & data valuation & 0.28 & 0.24 & 0.62 & 55 & 0.19 & \underline{0.78} & 35 & -1.9 & 0 \\
KNN-Shapley~\cite{jia2019efficient} & data valuation & \underline{0.63} & \underline{0.59} & \underline{0.76} & \underline{71} & \underline{0.44} & 0.73 & 26 & -0.2 & 0 \\
\midrule
\rowcolor{blue!8}\textbf{ADEM (ours)} & utility & \textbf{0.85} & \textbf{0.82} & \textbf{0.92} & \textbf{84} & \textbf{0.58} & \textbf{0.90} & \textbf{82} & \textbf{+2.6} & \textbf{46} \\
\bottomrule
\end{tabular}
\end{table*}

ADEM is best on all nine tests (\cref{tab:main}). It ranks candidate sets at  $\rho_{\text{all}}=0.85$, against 0.63 for KNN-Shapley, 0.53 for CAS and CHIPS, 0.45 for Cleanlab and at most 0.38 for any distance between distributions, and keeps $\rho_{\text{fix}}=0.82$ on fixed-size candidates, where FID drops to 0.20. It separates useful from other candidates at AUROC 0.92, moves in the right direction for 84\% of candidates (FID: 54\%), and its first choice delivers 82\% of the best achievable gain, against 44\% for the best baseline (CLIPScore) and about $0\%$ for FID, KID, FD-DINOv2, CMMD, Precision, Recall, OTDD and LAVA, whose first choice is the copy of $T$ on every test bed.

\begin{figure}[t]
  \centering
  \includegraphics[width=\columnwidth]{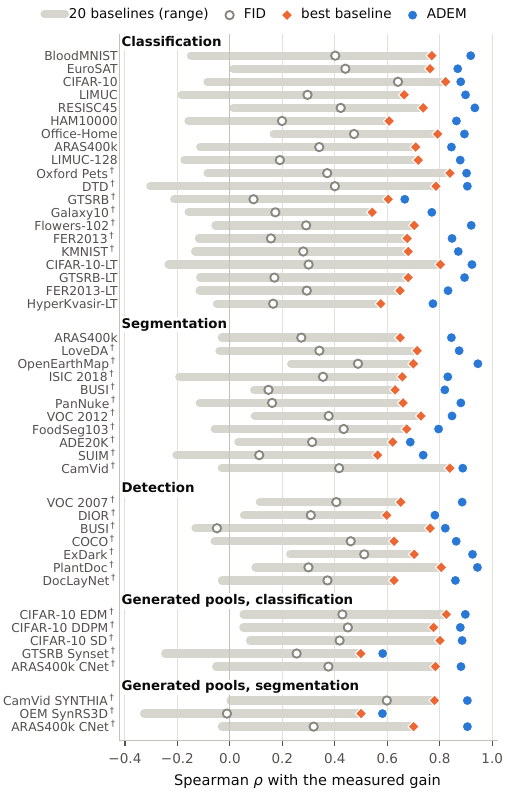}
  \caption{\textbf{Per-test-bed comparison.} Spearman correlation $\rho$ between each metric and the measured gain over all candidates of each test bed. $^\dagger$: held-out test bed.}
  \label{fig:perbed}
\end{figure}

ADEM ranks best on all 46 test beds (\cref{fig:perbed}). Its lowest $\rho$, 0.58, is on OpenEarthMap SynRS3D and GTSRB Synset, on every other test bed it is at least 0.67. ADEM is ahead of KNN-Shapley on every test bed with a one-sided Wilcoxon signed-rank test giving $p=1.4\times10^{-14}$, the smallest value attainable with 46 pairs. The advantage holds for every task as $\rho=\text{0.86, 0.82 and 0.87}$ on the classification, segmentation and detection test beds, against 0.64, 0.60 and 0.65 for KNN-Shapley. On the 32 held-out test beds ADEM reaches $\rho=0.84$, against 0.65 for KNN-Shapley and 0.31 for FID, and is best on all 32. This matches the design test beds ($\rho=0.87$), so the fixed settings transfer without tuning.

Fidelity metrics fail on two kinds of candidate. A copy of $T$ has $\text{FID}=0$ and is ranked first by FID, KID and FD-DINOv2 on all 46 test beds (CMMD: 44), yet it changes the task metric by only about $\pm1$ points while ADEM scores it close to zero as every copy has $u(d)=0$. Mislabeled sets keep the FID of clean data. For example, on Oxford Pets, the clean and half-mislabeled sets share FID 52.9 but change F1 by $+13.9$ and $-0.8$ points, while ADEM gives $+0.75$ and $-0.30$. As labels are corrupted, the gain falls at each step on all 46 test beds and ADEM on all of them, and the EM estimate recovers the share of correct labels of 220 sets with a mean absolute error of 0.032, without a single verified label in $D$.

\begin{figure}[t]
  \centering
  \includegraphics[width=\columnwidth]{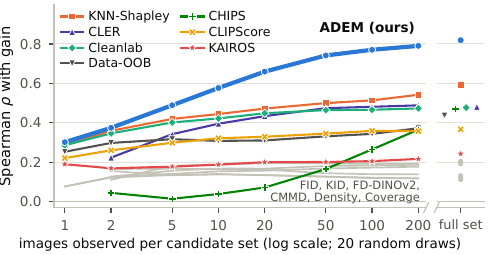}
  \caption{\textbf{Ranking candidate sets from few images.} Each fixed-size candidate is represented by $n$ random images (20 draws); Spearman $\rho$ with its gain, mean over the 46 test beds; full set: every image of the set ($\rho_\text{fix}$ in \cref{tab:main}).}
  \label{fig:smalln}
\end{figure}

With only 10 images per candidate, ADEM ranks at $\rho=0.58$ (classification 0.61, segmentation 0.53, detection 0.58), above KNN-Shapley with 200 images (0.54), while FID-type fidelity metrics stay near noise at every size (\cref{fig:smalln}). Every quantity in ADEM is a per-image sum, so there is no covariance to estimate and no small-sample bias~\cite{chong2020effectively}. Scoring images one at a time with $\phi(d)$ (\cref{sec:setscore}), ADEM reaches AUROC$_1$ 0.90 against 0.78 for KAIROS and 0.77 for Cleanlab (\cref{tab:main}).

On the eight generated pools, where every candidate set is built from a generator's images, ADEM ranks the candidate sets at $\rho=0.81$, against 0.65 for KNN-Shapley, 0.56 for CAS and 0.35 for FID, and is best on all eight. ADEM predicts with a negative score $-0.17$ that the rendered SynRS3D tiles lower OpenEarthMap's $-4.9$ F1 points, which no distance to $T$ can express.

ADEM can also be used to clean a corrupted pool of images by calculating $\phi(d)$ over a classification test bed of 200 clean, 200 copies, 200 mislabeled, 200 severely corrupted (uniform noise) and 200 task-irrelevant images. Keeping the images with $\phi(d)>0$ retains 77\% of the clean images, 0.3\% of the copies, 12\% of the mislabeled and 0.2\% of the corrupted and 1.2\% of the task-irrelevant images, providing +2.6 F1 score. The oracle of this setup, keeping only and all clean images, provides +3.8 F1 score while using all images loses 5.0 F1 score. KNN-Shapley (-0.2) and Cleanlab (+0.4) fall behind, keeping most of the corrupted or task-irrelevant images that cost F1, and LAVA (-2.5), CLIPScore (-0.1), realism (-3.3) and Density (-3.5) lose points as well. Data-OOB (+0.5) keeps every copy of $T$ but only 20\% of the clean images, CHIPS (-0.7) keeps 63\% of the copies but only 40\% of the clean images, and KAIROS (-1.9) keeps 66\% of the mislabeled images.

For the ablation studies, we changed one part of ADEM at a time and our results indicate that each part of ADEM contributes to the metric. Dropping uniqueness harms ADEM the most ($\rho$ 0.85$\to$0.71), followed by fixing  $\lambda=\frac12$ instead of estimating it (0.79), a $k$-NN label model (0.80), using the raw views only (0.82), CLIP alone (0.82), dropping task relevance (0.82). Two variants were found in the ablation to score higher, by at most 0.002, but we did not adopt them as it would be tuning on the held-out test beds.

%% file: sec/7_conclusion.tex
\section{Conclusion}
\label{sec:conclusion}
Whether new data will improve a model is a question of utility, not fidelity. Without training, ADEM estimates it by scoring each new image against the trusted set for task relevance, label correctness and uniqueness. Across 1151 candidate sets with measured gains from 46 classification, segmentation, and detection tasks, ADEM predicts which data will improve performance more accurately than existing metrics. It generalizes to held-out tasks without tuning, remains effective with long-tailed trusted sets, and works even for a small set.

The evaluations also show why existing metrics fall short as fidelity metrics ignore labels and rank copies of $T$ first, while label-quality and data-valuation methods rate copies of $T$ above clean images. By scoring each image against $T$ on all three properties, ADEM provides a meaningful zero point, so the same score can rank candidate sets, clean a pool and assess an individual image. 

ADEM's scores rank candidate sets reliably ($\rho$ = 0.85), and their sign agrees with candidate usefulness for 82\% of candidates. The remaining disagreements are expected from fixed settings. The costs $\theta$ and $\mu$ are shared by all test beds, whereas the harm of a wrong label depends on the learner, as models often tolerate moderate label noise~\cite{si2026can}. Likewise, an image transform is judged the same way on every task, although its effect is task-dependent. Our study is also limited to small trusted sets, frozen general-purpose encoders, and fixed training budgets. Measuring the gains takes 3591 trainings (three seeds for each of the 1151 candidate and 46 trusted sets on every test bed). Natural extensions include adapting the costs to the learner, using masks and boxes directly rather than reducing each image to one class, and calibrating the label probe to the style of generated images.

%% file: sec/X_suppl.tex
\clearpage
\maketitlesupplementary
\appendix
\crefalias{section}{appendix}

\section{Implementation Details of ADEM}
\label{app:impl}

We apply a Gaussian low-pass filter with standard deviation $W/32$ and kernel size $2\lceil3W/32\rceil+1$ where $W$ is the image width. We resize all images to $224\times224$ with bicubic interpolation and normalize them with each encoder's channel means and standard deviations. We then fit a multinomial logistic regression to the CLIP and DINOv2 embeddings of the low-pass filtered trusted image. We select regularization strength from $\{0.3,1,3,10,30\}$ and temperature from 90 values log-spaced in $[10^{-3},30]$ with mean out-of-fold log-likelihood of stratified cross-validation on $T$ with 5 folds. For trusted sets with fewer than five classes, we set the number of cross-validation folds equal to the number of classes. We then fit the logistic regression model on all of $T$ using the selected regularization strength and temperature. The label correctness prior~\eqref{eq:prior} fit the temperature and $\lambda_T$ on the out-of-fold probabilities of the trusted images. The kNN label model in the ablations is used without this prior.

The expectation-maximization (EM) estimate of $\lambda$~\eqref{eq:em} is initialized at $\lambda=\frac12$ and terminates after 500 iterations or when the change in $\lambda$ falls below $10^{-7}$. $\lambda_T$ is the same estimate with $r(d)=1$ on the out-of-fold probabilities. A candidate with a label different than the labels of $T$ gets $l(d)=0$, thus $q(d)=0$ and $u(d)=0$. In the set score, such candidates are treated as incorrectly labeled. ADEM also supports single-image scoring and pool cleaning. For an individual image, we compute $\phi(d)$ using $\lambda_T$ and set $m(d)=1$. For pool cleaning, we treat the whole pool as $D$, so that $\lambda$ is estimated on the pool and retain the images with $\phi(d)>0$.

\section{Benchmark Details}
\label{app:benchmark}

Images in classification benchmark are center-cropped to a square and resized with bicubic interpolation to $32\times32$ pixels for the test beds trained from scratch and to $128\times128$ for the fine-tuning. Datasets GTSRB, Galaxy10, Flowers-102, FER2013, KMNIST and DTD are randomly subsampled to at most 8000 training and 4000 test images. Office-Home dataset use real domain as the task and other three domains as candidates. Tiles of the ARAS400k dataset are labeled according to the dominant land-cover class. Synthetic counterparts of LIMUC and ARAS400k are generated by Generative Adversarial Networks~\cite{caglar2023ulcerative} and ControlNet~\cite{zhang2023adding} conditioned on land-cover maps. 3000 synthetic images, 300 per class labeled according to the generated classes, for CIFAR-10 were sampled from 1M EDM~\cite{karras2022elucidating}, another 3000 from 1M DDPM~\cite{ho2020denoising} samples, labeled by a CIFAR-10 classifier, finally, another 3000 images from CIFAKE~\cite{bird2024cifake}, generated by Stable Diffusion 1.4~\cite{rombach2022high} with labels from class prompts. For GTSRB dataset, 3000 synthetic images of Synset SignSet Germany dataset~\cite{sielemann2024synset} with same GTSRB's class IDs were used. For long-tail experiments CIFAR-10-LT, GTSRB-LT, HyperKvasir-LT and FER2013-LT, we reused images, test split learner and task-irrelevant images of CIFAR-10, GTSRB, HyperKvasir and FER2013.

For segmentation experiments, images are center-cropped to a square and resized to $128\times128$ with bicubic interpolation and corresponding segmentation masks with nearest-neighbor. Larger images were sampled as tiles and resized to $128\times128$. ARAS400k~\cite{caglar2026grounding} consists of Sentinel-2 tiles with ESA WorldCover segmentation masks and their synthetic counterparts generated with ControlNet. Tiles of Europe, Korea and California-Nevada from BELDE~\cite{caglar2026belde}, with the same classes, serve as other-region candidates. Rural and urban images of LoveDA~\cite{wang2021loveda} is used for trusted and candidate sets respectively. For OpenEarthMap~\cite{xia2023openearthmap}, we keep 30\% of its regions as other-region candidates. Task-1 of the ISIC 2018~\cite{codella2019skin} dataset for skin lesions and BUSI~\cite{aldhabyani2020dataset} dataset for benign or malignant segmentation of scans are used in their respective experimentation beds. PanNuke~\cite{gamper2019pannuke} uses nucleus instances of breast tissue as semantic masks of five nucleus types with colon and other 17 tissues as candidate sets with folds 1 and 2 for training and 3 for testing. For VOC 2012~\cite{pascal-voc-2012}, we use their official training and validation splits with empty boundaries converted to background. For food and ingredient segmentation we use FoodSeg103~\cite{wu2021large} with 103 ingredients and their backgrounds, and ADE20K~\cite{zhou2017scene} for scene understanding with 150 scene classes. For underwater imagery, SUIM~\cite{islam2020semantic} contributes with eight classes for training and test sets. With their 11 classes, we divide CamVid~\cite{brostow2009semantic} driving video frames into tiles and create pools from their training and validation frames and tests from their test frames. Synthetic candidates is prepared from 500 random frames of SYNTHIA-Rand~\cite{ros2016synthia}, rendered at the same resolution and divided with matching classes. OpenEarthMap synthetic counterpart with 500 images with $512\times512$ resolution of SynRS3D~\cite{song2024synrs3d} are also divided into tiles with matching classes as well. For segmentation tests, we compute an image's class with the mask at $64\times64$ resolution. 

To prepare detection experiments, images are scaled down with letter boxing width or height whichever is longer scaling down to 256 pixels with gray padding, preserving their aspect ratio. VOC 2007~\cite{pascal-voc-2007} supplies 3600 train and validation images and 1000 test images while we discarded boxes specifically marked as "difficult". For DIOR~\cite{li2020object}, we extract each scene at 800 pixel resolution and divide it in four tiles before letter boxing, with training and test sets set before division. NWPU VHR-10~\cite{cheng2014multi}, with matching ten classes and 3600 images, is selected as a candidate set for the DIOR dataset. Similarly, COCO~\cite{lin2014microsoft} with 80 classes and 3600 images for training and 1000 for test from the respective training and validation sets is prepared. DocLayNet~\cite{pfitzmann2022doclaynet} with 11 layout classes of document pages also provides same number of training and test images. ExDark~\cite{loh2019getting} and PlantDoc~\cite{singh2020plantdoc} with 12 low light and 27 different leaf classes respectively.

We draw a trusted set and a candidate pool from the available images for every test bed. On the segmentation and detection tests, some classes are the main class for only a few images thus some had to be unbalanced in classes. On FoodSeg103, ADE20K, SUIM, CamVid, COCO, ExDark, PlantDoc and DocLayNet, a class that is the main class of fewer than two images is left out of $T$. The candidate pool consist of up to 3000 randomly ordered remaining training images. Clean, transformed and relabeled candidates are selected from this pool. For long-tailed experiments, CIFAR-10-LT, GTSRB-LT and FER2013-LT, $T$ is about the same size as on the balanced test bed it comes from. On CIFAR-10-LT, class $i$ of $C$ gets a share proportional to $r^{-i/(C-1)}$ with imbalance ratio $r=100$~\cite{cao2019learning}, while GTSRB-LT, FER2013-LT and HyperKvasir-LT keep their natural class frequencies, with at least two images per class. Their pools hold up to 3000 of the other images in the same proportions, so their clean candidates are long-tailed as well.

The candidate sets are formed per test bed with clean draws of 100, 250, 500, 1000 and 2000 images and two more independent 500 image draws. For duplicates a copied, rotated, flipped or repeated images of the existing trusted set are used. For the label noise, we changed the first pool's images with 25, 50 or 100\% of their labels. We also used the worst human annotations of CIFAR-10N~\cite{wei2022learning}. For data imbalance, a geometric long tail were created for existing datasets or long-tail variants of them were used. Transforms preserving the content include rotation, flip, blur, compression and mild noise artifacts, while transforms that corrupt the content include heavy noise, pixelation, patch shuffling. Task-irrelevancy is obtained by selecting images from another test bed, either by mixing together with relevant (in-distribution) data or by themselves. For generative data, we samples from the publicly available synthetic sets for CIFAR-10, CIFAR-10-LT, GTSRB, GTSRB-LT, LIMUC, LIMUC-128, ARAS400k, CamVid and OpenEarthMap. For segmentation and detection test beds, label noise swaps masks between images or replaces other classes at random, and horizontal flip is used instead of rotation and flip for the detection tests.

Rotation and flip is defined as a random multiple of $90^\circ$ rotation then a horizontal flip with 50\% probability. Gaussian blur is defined by the standard deviation $0.7W/32$ where $W$ is the width of the image and JPEG compression quality is selected as 40. Mild and heavy noise is defined by additive Gaussian with standard deviations of 8 and 100 respectively. Pixelation is achieved by downsampling to $4\times4$ and upsampling with nearest-neighbor and patch shuffle randomly shuffles patches of a $4\times4$ grid. Segmentation masks and detection boxes follow the transformations.

We use the same learners for the same test beds to compare different methods fairly. For the $32\times32$ pixel resolution, we use CNN with four blocks of $3\times3$ convolution, batch normalization and ReLU activation function with 32, 64, 128 and 256 channels. Max-pooling after the first three layers and global average pooling and a linear layer at the end. Learning rate is $2*10^{-3}$ and weight decay is $5*10^{-4}$ for the trainings. For $128\times128$ resolution, we use ResNet-18 with a ImageNet-pretrained network and a new linear head, fine-tuned with learning rate of $5*10^{-4}$ and weight decay of $10^{-4}$. For segmentation tests, we use FPN (Feature Pyramid Network) with four stages of ImageNet pre-trained ResNet-18, lateral convolutions to 128 channels, merged top to down to stride 4 and followed by a $3\times3$ convolution with batch normalization and ReLU activation function followed by a $1\times1$ classifier head and bilinear upsampling with cross-entropy plus Dice loss as the loss function with learning rate and weight decay of $5*10^{-4}$ and $10^{-4}$, respectively. For the detection runs, we used TorchVision's MobileNetV3-Large FPN detector in the $320\times320$ pixel configuration with an ImageNet backbone. Learning rate $2*10^{-4}$ and weight decay of $10^{-4}$ are used for the fine-tuning. Classification's Macro-F1 is measured on the test split, segmentation's pixel macro-F1 is computed from the confusion matrix of all test pixels and averaged over the classes present in the test split and detection's mAP@0.5 is the mean over classes interpolated average precision at IoU$\ge0.5$.

\section{Baseline Implementations}
\label{app:baselines}

For a fair comparison, all label-aware baselines use the same primary class per image (\cref{sec:problem}). FID, KID and IS use Inception-v3 features and logits (torch-fidelity) and inception-v3-compat with $T$ as the reference, KID use unbiased MMD$^2$ with the cubic polynomial kernel and IS is computed on the whole set. FD-DINOv2 is the Fréchet distance on the raw-view DINOv2 ViT-S/14 embeddings and CMMD is the unbiased MMD$^2$ with a Gaussian kernel ($\sigma=10$) on raw-view CLIP ViT-B/32 embeddings. Both use the same encoders as ADEM while the original papers use larger ones, so that the comparison reflects the metrics rather than the encoders. Precision, recall, density and coverage use Inception features and $k=5$ nearest-neighbor clusters and the per-image realism score uses $k=3$. Vendi is the exponential of the entropy of the normalized eigenvalues of the similarity matrix of the representations. OTDD use squared orthogonal feature costs and square of the Wasserstein distance between diagonal Gaussians fitted per class, solved by log-domain Sinkhorn with $\epsilon=0.05$ mean cost. LAVA scores the candidate images with entropic optimal transport ($\epsilon=0.1$) with OTDD's label-aware cost, using the same diagonal Gaussians, a recent sliced variant reduces this to near-linear cost~\cite{nguyen2025lightspeed}. A set score is the unbiased Sinkhorn divergence according to the $T$. KNN-Shapley calculates Shapley values of the candidate images for a 10-nearest-neighbor classifier on the combined set of $T$ and $D$ with the trusted images as leave-one-out test points~\cite{jia2019efficient}. CLIPScore is the cosine similarity between raw-view CLIP image embedding and the basic description of "a photo of a (CLASS)" and its set score is calculated by the mean of all images. Cleanlab self confidence is the probability of a logistic regression trained on the feature space of $T$ assigning candidate's label with the set score being the mean of all images. Finally, CAS fits a logistic regression on the candidate set and reports the macro-F1 on the trusted-set, thus a candidate set with a single class predicts its class for every trusted image. CLER~\cite{li2025benchmarking} takes the mean raw-view CLIP ViT-B/32 image embedding of each class of the candidate set as its center and reports the accuracy of assigning each trusted image to the nearest center. Data-OOB~\cite{kwon2023dataoob} fits a random forest of 800 trees on $T$ and $D$ together and values a candidate image by how often the trees that did not see it predict its label. CHIPS~\cite{zhuang2026chips} scores each image, paired with the prompt of its class, with its CLIP contrastive-loss gradient on the mean gradient of $T$, times its learnability and its relevance to $T$ ($\beta=0.5$). With one image the loss has no negative, so its curve in \cref{fig:smalln} starts at two images. KAIROS~\cite{zhu2026kairos} values an image by its influence on the maximum mean discrepancy to $T$ (Gaussian kernel, median bandwidth) subtracting a residual of its logistic regression trained on $T$. The set score of each per-image method is the mean or the sum of its values, whichever ranks better on the 14 development test beds. OTDD, LAVA, KNN-Shapley, Cleanlab, CAS, Data-OOB, KAIROS, Vendi and ADEM use the same representation space (same encoders).

\section{Additional Results}
\label{app:extra}

Every classification test bed with a balanced $T$ gets a pool of clean, mislabeled (another class at random), severely corrupted (uniform random noise), task-irrelevant (another test bed's images with random labels) and copies of $T$, with 200 images each. Every pool is trained with a CNN from scratch at $32\times32$ resolution on all 21 test beds. Every score has its own threshold, for example ADEM uses $\phi(d)>0$, KNN-Shapley and LAVA use values $>0$, Cleanlab use confident learning~\cite{northcutt2021confident}, CLIPScore keeps images with a cosine $\geq0.28$, Precision~\cite{kynkaanniemi2019improved} use the realism score $\geq1$, density keep samples inside a kNN cluster of $T$, Data-OOB drops the lower of two $k$-means clusters of its values~\cite{kwon2023dataoob}, CHIPS keeps its top 30\%~\cite{zhuang2026chips} and KAIROS keeps the images with a positive value~\cite{zhu2026kairos}. Adding 200 corrupted or 200 task-irrelevant images to 200 clean images lowers F1 score on 11 and 14 of the 21 test beds respectively. ADEM keeps 77\% of the clean images, 12\% of the mislabeled and 0.2\% of the corrupted and 1.2\% of the task-irrelevant images.

\begin{figure*}[!t]
  \centering
  \includegraphics[width=\textwidth]{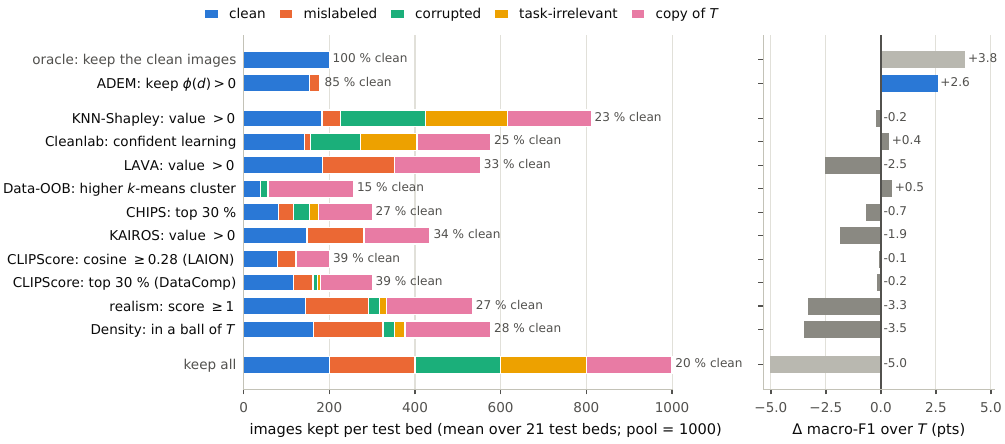}
  \caption{Cleaning a corrupted pool of 1000 images per test bed (200 each of clean, mislabeled, corrupted, task-irrelevant and copies of $T$ images) on all 21 classification test beds. Left: how many images each rule keeps (mean per test bed) and the share of it that is clean. ADEM keeps the images with $\phi(d)>0$; every baseline keeps the images that pass its own threshold. Right: macro-F1 gain of training on $T$ plus the kept images.}
  \label{fig:pool}
\end{figure*}

The ablations of the experimental setup of \cref{sec:results} are detailed in \Cref{tab:ablation} and \Cref{tab:perbed} shows correlation of every metric with the measured performance on every test bed.

\begin{table}[ht]
\centering
\footnotesize
\setlength{\tabcolsep}{2.6pt}
\caption{The ablations with changes to the infrastructure of ADEM with means over the 46 test beds and per task.}
\label{tab:ablation}
\begin{tabular}{@{}l cc ccc@{}}
\toprule
 & \multicolumn{2}{c}{All 46 test beds} & Cls. & Seg. & Det. \\
\cmidrule(lr){2-3}\cmidrule(l){4-6}
Variant & $\rho_{\text{all}}$ & $\rho_{\text{fix}}$ & $\rho_{\text{all}}$ & $\rho_{\text{all}}$ & $\rho_{\text{all}}$ \\
\midrule
\rowcolor{blue!8}ADEM (default) & 0.848 & 0.819 & 0.856 & 0.824 & 0.868 \\
\midrule
no prior correction & 0.828 & 0.795 & 0.821 & 0.820 & 0.871 \\
no uniqueness ($u(d)=1$) & 0.715 & 0.673 & 0.744 & 0.672 & 0.697 \\
no EM ($\lambda=\frac12$) & 0.788 & 0.760 & 0.795 & 0.796 & 0.749 \\
no task relevance ($r{=}1$, $\mu{=}0$) & 0.825 & 0.791 & 0.832 & 0.787 & 0.878 \\
raw views only & 0.816 & 0.782 & 0.844 & 0.785 & 0.781 \\
low-pass filtered only & 0.848 & 0.821 & 0.861 & 0.805 & 0.886 \\
CLIP only & 0.818 & 0.788 & 0.827 & 0.778 & 0.865 \\
DINOv2 only & 0.846 & 0.817 & 0.847 & 0.827 & 0.879 \\
probe on all four views & 0.836 & 0.808 & 0.855 & 0.813 & 0.817 \\
kNN label model & 0.803 & 0.772 & 0.814 & 0.755 & 0.862 \\
\midrule
$k=5$ & 0.846 & 0.818 & 0.855 & 0.820 & 0.867 \\
$k=20$ & 0.849 & 0.819 & 0.858 & 0.824 & 0.865 \\
$\alpha=0.02$ & 0.850 & 0.823 & 0.859 & 0.822 & 0.877 \\
$\alpha=0.1$ & 0.841 & 0.812 & 0.849 & 0.811 & 0.872 \\
$\theta=2$ & 0.834 & 0.802 & 0.850 & 0.788 & 0.864 \\
$\theta=5$ & 0.842 & 0.816 & 0.836 & 0.846 & 0.854 \\
$\theta=10$ & 0.798 & 0.775 & 0.794 & 0.796 & 0.815 \\
$\mu=1$ & 0.828 & 0.798 & 0.834 & 0.802 & 0.861 \\
$\mu=2$ & 0.796 & 0.761 & 0.801 & 0.761 & 0.848 \\
$h=\frac1{10}$ & 0.829 & 0.806 & 0.850 & 0.784 & 0.846 \\
$h=\frac12$ & 0.823 & 0.799 & 0.826 & 0.809 & 0.843 \\
\bottomrule
\end{tabular}
\end{table}

\begin{table*}[!t]
\centering
\scriptsize
\setlength{\tabcolsep}{1.5pt}
\caption{Spearman $\rho$ with the measured gain, per test bed (all candidates and every metric oriented so that higher means better data). Best per test bed in bold. $^\ast$: CNN trained from scratch. $^\dagger$: held-out test bed.}
\label{tab:perbed}
\begin{tabular}{@{}l *{20}{c} >{\columncolor{blue!8}}c@{}}
\toprule
Bed & FID & KID & FD-D. & CMMD & IS & Prec. & Dens. & Rec. & Cov. & Vendi & OTDD & LAVA & CLIP-S & CHIPS & CAS & CLER & Clean. & D-OOB & KAIR. & KNN-Sh. & ADEM \\
\midrule
\multicolumn{22}{@{}l}{\emph{Classification}} \\
BloodMNIST$^\ast$ & 0.40 & 0.14 & 0.41 & 0.23 & -0.15 & 0.43 & 0.39 & 0.32 & 0.49 & -0.11 & 0.57 & 0.56 & 0.59 & 0.36 & 0.61 & 0.65 & 0.69 & 0.65 & 0.09 & 0.77 & \textbf{0.92} \\
EuroSAT$^\ast$ & 0.44 & 0.47 & 0.39 & 0.42 & 0.31 & 0.01 & 0.07 & 0.28 & 0.42 & 0.43 & 0.56 & 0.52 & 0.38 & 0.55 & 0.64 & 0.71 & 0.57 & 0.57 & 0.53 & 0.76 & \textbf{0.87} \\
CIFAR-10$^\ast$ & 0.64 & 0.21 & 0.59 & 0.21 & 0.30 & 0.43 & 0.54 & -0.09 & 0.68 & 0.49 & 0.75 & 0.75 & 0.50 & 0.66 & 0.79 & 0.75 & 0.64 & 0.65 & 0.50 & 0.82 & \textbf{0.88} \\
LIMUC$^\ast$ & 0.30 & 0.05 & 0.32 & 0.22 & 0.15 & 0.31 & 0.06 & 0.29 & 0.31 & 0.33 & 0.22 & 0.25 & -0.18 & 0.66 & 0.12 & 0.43 & 0.48 & 0.33 & 0.44 & 0.59 & \textbf{0.90} \\
RESISC45 & 0.42 & 0.06 & 0.34 & 0.01 & 0.33 & 0.10 & 0.25 & 0.22 & 0.31 & 0.42 & 0.50 & 0.47 & 0.45 & 0.71 & 0.66 & 0.62 & 0.49 & 0.50 & 0.23 & 0.74 & \textbf{0.93} \\
HAM10000 & 0.20 & -0.06 & 0.20 & -0.16 & 0.28 & 0.07 & 0.04 & -0.03 & 0.25 & 0.37 & 0.08 & 0.13 & 0.27 & 0.61 & 0.14 & 0.46 & 0.36 & 0.54 & 0.05 & 0.54 & \textbf{0.86} \\
Office-Home & 0.47 & 0.43 & 0.52 & 0.32 & 0.51 & 0.17 & 0.24 & 0.33 & 0.53 & 0.62 & 0.59 & 0.56 & 0.56 & 0.46 & 0.60 & 0.60 & 0.38 & 0.48 & 0.46 & 0.79 & \textbf{0.89} \\
ARAS400k & 0.34 & 0.22 & 0.29 & 0.28 & -0.11 & 0.24 & 0.23 & 0.19 & 0.46 & 0.22 & 0.41 & 0.42 & -0.11 & 0.70 & 0.58 & 0.59 & 0.39 & 0.59 & 0.21 & 0.71 & \textbf{0.84} \\
LIMUC-128 & 0.19 & 0.12 & 0.30 & 0.23 & 0.01 & 0.07 & -0.12 & 0.30 & 0.15 & 0.36 & 0.23 & 0.24 & -0.17 & 0.46 & 0.24 & 0.34 & 0.63 & 0.38 & 0.34 & 0.72 & \textbf{0.88} \\
Oxford Pets$^\dagger$ & 0.37 & 0.14 & 0.35 & 0.05 & 0.56 & -0.02 & 0.12 & -0.09 & 0.37 & 0.63 & 0.53 & 0.53 & 0.39 & 0.46 & 0.66 & 0.60 & 0.44 & 0.46 & 0.42 & 0.84 & \textbf{0.90} \\
DTD$^\dagger$ & 0.40 & 0.25 & 0.44 & 0.25 & 0.33 & -0.30 & 0.27 & 0.36 & 0.48 & 0.39 & 0.60 & 0.56 & 0.49 & 0.75 & 0.74 & 0.67 & 0.57 & 0.43 & 0.45 & 0.79 & \textbf{0.91} \\
GTSRB$^\dagger$ & 0.09 & 0.00 & 0.12 & 0.08 & 0.16 & -0.12 & -0.21 & 0.00 & 0.16 & 0.51 & 0.09 & 0.16 & 0.30 & 0.60 & 0.55 & 0.51 & 0.13 & 0.51 & 0.15 & 0.46 & \textbf{0.67} \\
Galaxy10$^\dagger$ & 0.17 & -0.10 & 0.07 & -0.06 & 0.53 & -0.01 & -0.16 & 0.00 & 0.36 & 0.32 & 0.27 & 0.26 & 0.07 & 0.46 & 0.27 & 0.37 & 0.42 & 0.31 & 0.21 & 0.54 & \textbf{0.77} \\
Flowers-102$^\dagger$ & 0.29 & -0.05 & 0.33 & 0.21 & -0.02 & 0.15 & 0.16 & 0.29 & 0.26 & 0.45 & 0.52 & 0.52 & 0.36 & 0.64 & 0.64 & 0.54 & 0.33 & 0.46 & 0.27 & 0.70 & \textbf{0.92} \\
FER2013$^\dagger$ & 0.16 & 0.00 & 0.27 & 0.13 & 0.12 & -0.12 & -0.07 & 0.23 & 0.24 & 0.42 & 0.23 & 0.15 & 0.14 & 0.53 & 0.45 & 0.28 & 0.34 & 0.46 & 0.33 & 0.68 & \textbf{0.85} \\
KMNIST$^\ast$$^\dagger$ & 0.28 & 0.05 & 0.27 & 0.18 & -0.13 & 0.22 & 0.23 & -0.10 & 0.23 & 0.06 & 0.36 & 0.32 & 0.09 & 0.15 & 0.67 & 0.59 & 0.68 & 0.56 & 0.20 & 0.67 & \textbf{0.87} \\
CIFAR-10-LT$^\ast$ & 0.30 & 0.13 & 0.13 & 0.06 & 0.80 & 0.63 & 0.26 & 0.06 & 0.25 & 0.53 & -0.05 & 0.08 & 0.70 & 0.34 & 0.73 & 0.47 & -0.23 & 0.44 & 0.14 & 0.12 & \textbf{0.92} \\
GTSRB-LT & 0.17 & -0.07 & 0.21 & 0.00 & 0.23 & -0.03 & 0.18 & 0.09 & 0.16 & 0.44 & 0.14 & 0.21 & 0.66 & 0.68 & 0.67 & 0.53 & -0.11 & 0.48 & 0.12 & 0.29 & \textbf{0.89} \\
FER2013-LT & 0.29 & 0.12 & 0.33 & 0.09 & -0.12 & 0.13 & -0.03 & 0.27 & 0.35 & 0.31 & 0.36 & 0.25 & 0.35 & 0.38 & 0.65 & 0.38 & 0.17 & 0.35 & 0.25 & 0.31 & \textbf{0.83} \\
HyperKvasir-LT & 0.17 & -0.04 & 0.07 & 0.01 & 0.15 & -0.05 & 0.02 & 0.09 & 0.19 & 0.33 & 0.23 & 0.22 & 0.05 & 0.58 & 0.50 & 0.01 & -0.02 & 0.30 & 0.11 & 0.45 & \textbf{0.77} \\
\emph{mean} & 0.30 & 0.10 & 0.30 & 0.14 & 0.21 & 0.12 & 0.12 & 0.15 & 0.33 & 0.38 & 0.36 & 0.36 & 0.29 & 0.54 & 0.55 & 0.51 & 0.37 & 0.47 & 0.27 & 0.61 & \textbf{0.86} \\
\midrule
\multicolumn{22}{@{}l}{\emph{Segmentation}} \\
ARAS400k & 0.27 & 0.27 & 0.31 & 0.15 & 0.18 & -0.03 & -0.02 & 0.26 & 0.41 & 0.16 & 0.24 & 0.28 & 0.13 & 0.50 & 0.47 & 0.29 & 0.46 & 0.65 & 0.09 & 0.63 & \textbf{0.84} \\
LoveDA$^\dagger$ & 0.34 & 0.15 & 0.36 & 0.03 & 0.10 & -0.04 & 0.31 & 0.20 & 0.39 & 0.26 & 0.24 & 0.30 & 0.54 & 0.71 & 0.56 & 0.54 & 0.54 & 0.51 & 0.16 & 0.50 & \textbf{0.87} \\
OpenEarthMap$^\dagger$ & 0.49 & 0.38 & 0.49 & 0.39 & 0.24 & 0.23 & 0.47 & 0.33 & 0.56 & 0.44 & 0.53 & 0.50 & 0.69 & 0.43 & 0.64 & 0.68 & 0.70 & 0.33 & 0.46 & 0.67 & \textbf{0.95} \\
ISIC 2018$^\dagger$ & 0.36 & 0.18 & 0.31 & 0.32 & 0.10 & 0.28 & 0.31 & 0.16 & 0.23 & 0.27 & 0.36 & 0.35 & -0.19 & 0.36 & 0.44 & 0.66 & 0.36 & 0.28 & 0.31 & 0.57 & \textbf{0.83} \\
BUSI$^\dagger$ & 0.15 & 0.20 & 0.19 & 0.17 & 0.13 & 0.12 & 0.13 & 0.21 & 0.25 & 0.09 & 0.18 & 0.18 & 0.49 & 0.39 & 0.32 & 0.47 & 0.33 & 0.41 & 0.33 & 0.63 & \textbf{0.82} \\
PanNuke$^\dagger$ & 0.16 & 0.00 & 0.18 & 0.12 & 0.05 & -0.04 & -0.11 & 0.23 & 0.18 & -0.03 & 0.28 & 0.25 & 0.26 & 0.63 & 0.52 & 0.16 & 0.42 & 0.35 & -0.03 & 0.66 & \textbf{0.88} \\
VOC 2012$^\dagger$ & 0.38 & 0.28 & 0.33 & 0.15 & 0.31 & 0.10 & 0.26 & 0.22 & 0.27 & 0.22 & 0.44 & 0.41 & 0.61 & 0.54 & 0.60 & 0.60 & 0.73 & 0.51 & 0.42 & 0.71 & \textbf{0.85} \\
FoodSeg103$^\dagger$ & 0.43 & 0.00 & 0.39 & 0.07 & 0.01 & 0.35 & 0.46 & -0.06 & 0.37 & 0.06 & 0.46 & 0.42 & 0.40 & 0.67 & 0.46 & 0.67 & 0.51 & 0.64 & -0.04 & 0.63 & \textbf{0.80} \\
ADE20K$^\dagger$ & 0.31 & 0.31 & 0.36 & 0.23 & 0.24 & 0.06 & 0.05 & 0.21 & 0.40 & 0.35 & 0.43 & 0.45 & 0.03 & 0.62 & 0.35 & 0.54 & 0.50 & 0.45 & 0.29 & 0.59 & \textbf{0.69} \\
SUIM$^\dagger$ & 0.11 & 0.01 & 0.11 & 0.00 & -0.20 & 0.10 & 0.19 & 0.08 & 0.17 & -0.12 & 0.09 & 0.11 & 0.56 & 0.28 & 0.48 & 0.36 & 0.54 & 0.29 & 0.05 & 0.44 & \textbf{0.74} \\
CamVid$^\dagger$ & 0.42 & 0.28 & 0.39 & 0.26 & -0.03 & 0.25 & 0.38 & 0.17 & 0.36 & 0.02 & 0.51 & 0.52 & 0.71 & 0.63 & 0.43 & 0.41 & 0.71 & 0.60 & 0.32 & 0.84 & \textbf{0.89} \\
\emph{mean} & 0.31 & 0.19 & 0.31 & 0.17 & 0.10 & 0.12 & 0.22 & 0.18 & 0.33 & 0.16 & 0.34 & 0.34 & 0.38 & 0.52 & 0.48 & 0.49 & 0.53 & 0.46 & 0.21 & 0.63 & \textbf{0.83} \\
\midrule
\multicolumn{22}{@{}l}{\emph{Detection}} \\
VOC 2007$^\dagger$ & 0.41 & 0.33 & 0.38 & 0.31 & 0.51 & 0.39 & 0.30 & 0.11 & 0.43 & 0.40 & 0.57 & 0.50 & 0.25 & 0.58 & 0.59 & 0.59 & 0.59 & 0.52 & 0.32 & 0.65 & \textbf{0.89} \\
DIOR$^\dagger$ & 0.31 & 0.16 & 0.24 & 0.05 & 0.35 & 0.07 & 0.09 & 0.15 & 0.30 & 0.43 & 0.42 & 0.36 & 0.28 & 0.60 & 0.53 & 0.55 & 0.33 & 0.46 & 0.25 & 0.49 & \textbf{0.78} \\
BUSI$^\dagger$ & -0.05 & -0.03 & -0.05 & -0.11 & 0.13 & 0.10 & -0.10 & -0.13 & 0.03 & -0.06 & -0.04 & -0.04 & 0.46 & 0.64 & 0.19 & 0.29 & 0.41 & 0.58 & 0.01 & 0.76 & \textbf{0.82} \\
COCO$^\dagger$ & 0.46 & 0.18 & 0.45 & 0.37 & 0.55 & 0.25 & 0.20 & 0.37 & 0.52 & 0.43 & 0.59 & 0.59 & -0.06 & 0.58 & 0.53 & 0.63 & 0.41 & 0.27 & 0.03 & 0.57 & \textbf{0.86} \\
ExDark$^\dagger$ & 0.51 & 0.35 & 0.50 & 0.34 & 0.53 & 0.49 & 0.51 & 0.23 & 0.57 & 0.44 & 0.59 & 0.44 & 0.41 & 0.44 & 0.65 & 0.65 & 0.65 & 0.57 & 0.48 & 0.70 & \textbf{0.92} \\
PlantDoc$^\dagger$ & 0.30 & 0.15 & 0.30 & 0.10 & 0.25 & 0.30 & 0.29 & 0.28 & 0.28 & 0.30 & 0.54 & 0.48 & 0.52 & 0.63 & 0.50 & 0.59 & 0.53 & 0.48 & 0.53 & 0.81 & \textbf{0.94} \\
DocLayNet$^\dagger$ & 0.37 & 0.23 & 0.36 & 0.29 & -0.03 & 0.36 & 0.38 & 0.23 & 0.34 & 0.27 & 0.36 & 0.38 & 0.37 & 0.43 & 0.58 & 0.63 & 0.61 & 0.18 & 0.31 & 0.55 & \textbf{0.86} \\
\emph{mean} & 0.33 & 0.20 & 0.31 & 0.19 & 0.33 & 0.28 & 0.24 & 0.18 & 0.35 & 0.31 & 0.43 & 0.39 & 0.32 & 0.56 & 0.51 & 0.56 & 0.50 & 0.44 & 0.28 & 0.65 & \textbf{0.87} \\
\midrule
\multicolumn{22}{@{}l}{\emph{Generated pools}} \\
CIFAR-10 EDM$^\ast$$^\dagger$ & 0.43 & 0.26 & 0.49 & 0.40 & 0.44 & 0.30 & 0.25 & 0.05 & 0.42 & 0.42 & 0.59 & 0.61 & 0.75 & 0.49 & 0.66 & 0.67 & 0.56 & 0.66 & 0.57 & 0.83 & \textbf{0.90} \\
CIFAR-10 DDPM$^\ast$$^\dagger$ & 0.45 & 0.34 & 0.43 & 0.19 & 0.32 & 0.33 & 0.46 & 0.05 & 0.48 & 0.38 & 0.58 & 0.58 & 0.69 & 0.53 & 0.67 & 0.69 & 0.71 & 0.61 & 0.48 & 0.78 & \textbf{0.88} \\
CIFAR-10 SD$^\ast$$^\dagger$ & 0.42 & 0.28 & 0.49 & 0.22 & 0.44 & 0.23 & 0.45 & 0.08 & 0.45 & 0.41 & 0.48 & 0.45 & 0.71 & 0.56 & 0.56 & 0.64 & 0.75 & 0.62 & 0.58 & 0.80 & \textbf{0.89} \\
GTSRB Synset$^\dagger$ & 0.25 & 0.09 & 0.20 & 0.11 & 0.22 & -0.25 & -0.14 & 0.01 & 0.31 & 0.50 & 0.36 & 0.26 & 0.37 & 0.46 & 0.50 & 0.44 & 0.16 & 0.44 & 0.21 & 0.49 & \textbf{0.58} \\
ARAS400k CNet$^\dagger$ & 0.38 & 0.33 & 0.36 & 0.41 & 0.15 & 0.09 & 0.04 & 0.19 & 0.36 & 0.22 & 0.38 & 0.39 & -0.05 & 0.78 & 0.58 & 0.57 & 0.46 & 0.61 & 0.30 & 0.78 & \textbf{0.88} \\
CamVid SYNTHIA$^\dagger$ & 0.60 & 0.48 & 0.58 & 0.14 & 0.00 & 0.25 & 0.24 & 0.24 & 0.44 & 0.23 & 0.59 & 0.47 & 0.74 & 0.78 & 0.74 & 0.61 & 0.69 & 0.64 & 0.33 & 0.74 & \textbf{0.91} \\
OpenEarthMap SynRS3D$^\dagger$ & -0.01 & 0.44 & -0.02 & 0.31 & 0.10 & 0.19 & 0.07 & 0.30 & -0.05 & 0.21 & 0.06 & 0.03 & 0.50 & -0.33 & 0.30 & 0.31 & 0.40 & -0.06 & 0.45 & 0.14 & \textbf{0.58} \\
ARAS400k CNet seg.$^\dagger$ & 0.32 & 0.04 & 0.37 & 0.10 & -0.03 & 0.18 & 0.27 & 0.19 & 0.36 & 0.05 & 0.37 & 0.34 & 0.22 & 0.53 & 0.49 & 0.42 & 0.33 & 0.70 & -0.03 & 0.65 & \textbf{0.91} \\
\emph{mean} & 0.35 & 0.29 & 0.36 & 0.23 & 0.21 & 0.17 & 0.20 & 0.14 & 0.35 & 0.30 & 0.43 & 0.39 & 0.49 & 0.47 & 0.56 & 0.54 & 0.51 & 0.53 & 0.36 & 0.65 & \textbf{0.81} \\
\midrule
\emph{mean of all 46} & 0.32 & 0.17 & 0.31 & 0.17 & 0.20 & 0.15 & 0.18 & 0.16 & 0.34 & 0.30 & 0.38 & 0.37 & 0.35 & 0.53 & 0.53 & 0.52 & 0.45 & 0.47 & 0.28 & 0.63 & \textbf{0.85} \\
\bottomrule
\end{tabular}
\end{table*}

Synthetic and real candidate images and tests with them are detailed in \Cref{tab:synth}. ADEM ranks synthetic candidates in line with the measured gain. ADEM also successfully distinguishes the harm of synthetic candidates with a negative score.

\begin{table}[ht]
\centering
\footnotesize
\setlength{\tabcolsep}{3.2pt}
\caption{Real \vs synthetic candidates with 2000 images each (1000 on CamVid, whose pool holds fewer). $\Delta$: measured gain in points $\lambda$: ADEM's EM estimate of the share of correct labels. GAN and ControlNet are the synthetic images that come with LIMUC and ARAS400k. EDM, DDPM, Stable Diffusion (SD) and the rendered sets are the public synthetic data released.}
\label{tab:synth}
\begin{tabular}{@{}ll rrrr@{}}
\toprule
Bed & Source & $\Delta$ (pts) & FID$\downarrow$ & ADEM & $\lambda$ \\
\midrule
LIMUC & real & +10.6 & 53.9 & 1.00 & 0.995 \\
 & GAN & +2.8 & 55.8 & 0.09 & 0.674 \\
\addlinespace[2pt]
LIMUC-128 & real & +11.7 & 54.3 & 1.04 & 0.997 \\
 & GAN & +2.0 & 59.9 & 0.20 & 0.707 \\
\addlinespace[2pt]
ARAS400k & real & +5.6 & 66.4 & 1.31 & 0.987 \\
 & ControlNet & +3.2 & 74.7 & 1.20 & 0.975 \\
\addlinespace[2pt]
CIFAR-10 & real & +14.7 & 61.1 & 1.14 & 0.998 \\
 & EDM~\cite{wang2023better} & +17.7 & 61.5 & 1.07 & 0.999 \\
 & DDPM~\cite{gowal2021improving} & +16.6 & 61.5 & 1.05 & 0.998 \\
 & SD~\cite{bird2024cifake} & +5.1 & 87.5 & 0.91 & 0.999 \\
\addlinespace[2pt]
GTSRB & real & +35.0 & 52.5 & 1.25 & 0.999 \\
 & rendered~\cite{sielemann2024synset} & +32.6 & 135.3 & 0.49 & 0.725 \\
\addlinespace[2pt]
ARAS400k (seg.) & real & +7.0 & 76.8 & 0.81 & 0.997 \\
 & ControlNet & +9.6 & 75.4 & 0.93 & 0.998 \\
\addlinespace[2pt]
CamVid & real & +8.2 & 54.7 & 1.09 & 0.996 \\
 & rendered~\cite{ros2016synthia} & +2.4 & 170.3 & 0.15 & 0.970 \\
\addlinespace[2pt]
OpenEarthMap & real & +8.5 & 90.3 & 1.05 & 0.985 \\
 & rendered~\cite{song2024synrs3d} & -4.9 & 193.2 & -0.17 & 0.612 \\
\bottomrule
\end{tabular}
\end{table}

To show ADEM holds for larger trusted sets, we used $2\times$ and $3\times$ larger trusted sets and measured the gains of the candidate sets again. Not every dataset could provide these larger sets thus we used what was available and ADEM is the best metric both for $2T$ and $3T$ configurations compared to the other baselines, as highlighted in \cref{tab:largerT_beds}. 

\begin{table}[ht]
\centering
\scriptsize
\setlength{\tabcolsep}{4pt}
\caption{Larger trusted sets with $T$, $2T$ and $3T$ when applicable. Spearman $\rho$ with the measured gain for ADEM and for the best of the 20 baselines, per test bed, over all candidate sets except EDM, DDPM, SD and the rendered sets (\cref{tab:synth}), which were trained at $T$ only. Due to these exceptions, the $T$ column differs from \cref{tab:beds} on six test beds. $^\dagger$: held-out test bed.}
\label{tab:largerT_beds}
\begin{tabular}{@{}l ccc >{\columncolor{blue!8}}c>{\columncolor{blue!8}}c>{\columncolor{blue!8}}c ccc@{}}
\toprule
 & \multicolumn{3}{c}{$|T|$} & \multicolumn{3}{c}{ADEM $\rho$} & \multicolumn{3}{c}{best baseline $\rho$} \\
\cmidrule(lr){2-4}\cmidrule(lr){5-7}\cmidrule(l){8-10}
Test bed & $T$ & $2T$ & $3T$ & $T$ & $2T$ & $3T$ & $T$ & $2T$ & $3T$ \\
\midrule
\multicolumn{10}{@{}l}{\emph{Classification}} \\
BloodMNIST & 400 & 800 & 1200 & \textbf{0.92} & \textbf{0.85} & \textbf{0.83} & 0.77 & 0.78 & 0.75 \\
EuroSAT & 500 & 1000 & 1500 & \textbf{0.87} & \textbf{0.91} & \textbf{0.89} & 0.76 & 0.83 & 0.80 \\
CIFAR-10 & 500 & 1000 & 1500 & \textbf{0.87} & \textbf{0.88} & \textbf{0.86} & 0.79 & 0.78 & 0.75 \\
LIMUC & 400 & 800 & 1200 & \textbf{0.90} & \textbf{0.80} & \textbf{0.83} & 0.66 & 0.56 & 0.67 \\
RESISC45 & 450 & 900 & 1350 & \textbf{0.93} & \textbf{0.96} & \textbf{0.97} & 0.74 & 0.76 & 0.79 \\
HAM10000 & 210 & 418 & 587 & \textbf{0.86} & \textbf{0.84} & \textbf{0.69} & 0.61 & 0.58 & 0.57 \\
Office-Home & 325 & -- & -- & \textbf{0.89} & -- & -- & 0.79 & -- & -- \\
ARAS400k & 300 & 600 & 899 & \textbf{0.84} & \textbf{0.83} & \textbf{0.83} & 0.71 & 0.74 & 0.67 \\
LIMUC-128 & 400 & 800 & 1200 & \textbf{0.88} & \textbf{0.81} & \textbf{0.67} & 0.72 & 0.79 & 0.60 \\
Oxford Pets\smash{$^{\dagger}$} & 370 & 660 & -- & \textbf{0.90} & \textbf{0.88} & -- & 0.84 & 0.69 & -- \\
DTD\smash{$^{\dagger}$} & 470 & 759 & -- & \textbf{0.91} & \textbf{0.89} & -- & 0.79 & 0.76 & -- \\
GTSRB\smash{$^{\dagger}$} & 430 & 860 & 1284 & \textbf{0.79} & \textbf{0.70} & \textbf{0.63} & 0.56 & 0.47 & 0.40 \\
Galaxy10\smash{$^{\dagger}$} & 300 & 600 & 889 & \textbf{0.77} & \textbf{0.75} & \textbf{0.72} & 0.54 & 0.53 & 0.55 \\
Flowers-102\smash{$^{\dagger}$} & 510 & 1020 & 1530 & \textbf{0.92} & \textbf{0.92} & \textbf{0.93} & 0.70 & 0.72 & 0.72 \\
FER2013\smash{$^{\dagger}$} & 350 & 699 & 999 & \textbf{0.85} & \textbf{0.89} & \textbf{0.87} & 0.68 & 0.72 & 0.71 \\
KMNIST\smash{$^{\dagger}$} & 500 & 1000 & 1500 & \textbf{0.87} & \textbf{0.84} & \textbf{0.82} & 0.68 & 0.67 & 0.76 \\
CIFAR-10-LT & 499 & 998 & 1497 & \textbf{0.85} & \textbf{0.82} & \textbf{0.77} & 0.80 & 0.72 & 0.74 \\
GTSRB-LT & 431 & 851 & 1266 & \textbf{0.88} & \textbf{0.83} & \textbf{0.81} & 0.65 & 0.65 & 0.55 \\
FER2013-LT & 349 & 693 & 1037 & \textbf{0.83} & \textbf{0.86} & \textbf{0.85} & 0.65 & 0.48 & 0.54 \\
HyperKvasir-LT & 241 & 468 & 695 & \textbf{0.77} & \textbf{0.76} & \textbf{0.68} & 0.58 & 0.38 & 0.35 \\
\midrule
\multicolumn{10}{@{}l}{\emph{Segmentation}} \\
ARAS400k & 255 & 394 & 523 & \textbf{0.84} & \textbf{0.83} & \textbf{0.81} & 0.65 & 0.66 & 0.71 \\
LoveDA\smash{$^{\dagger}$} & 301 & 487 & 616 & \textbf{0.87} & \textbf{0.86} & \textbf{0.89} & 0.71 & 0.68 & 0.72 \\
OpenEarthMap\smash{$^{\dagger}$} & 310 & 526 & 701 & \textbf{0.95} & \textbf{0.95} & \textbf{0.93} & 0.72 & 0.77 & 0.74 \\
ISIC 2018\smash{$^{\dagger}$} & 200 & 349 & 449 & \textbf{0.83} & \textbf{0.87} & \textbf{0.82} & 0.66 & 0.65 & 0.58 \\
BUSI\smash{$^{\dagger}$} & 120 & 124 & -- & \textbf{0.82} & \textbf{0.71} & -- & 0.63 & 0.51 & -- \\
PanNuke\smash{$^{\dagger}$} & 223 & 381 & 473 & \textbf{0.88} & \textbf{0.81} & \textbf{0.79} & 0.66 & 0.59 & 0.59 \\
VOC 2012\smash{$^{\dagger}$} & 406 & 446 & -- & \textbf{0.85} & \textbf{0.79} & -- & 0.73 & 0.69 & -- \\
FoodSeg103\smash{$^{\dagger}$} & 333 & 547 & 686 & \textbf{0.80} & \textbf{0.90} & \textbf{0.88} & 0.67 & 0.74 & 0.74 \\
ADE20K\smash{$^{\dagger}$} & 244 & 364 & 441 & \textbf{0.69} & \textbf{0.77} & \textbf{0.78} & 0.62 & 0.67 & 0.69 \\
SUIM\smash{$^{\dagger}$} & 217 & 377 & 480 & \textbf{0.74} & \textbf{0.77} & \textbf{0.74} & 0.56 & 0.67 & 0.70 \\
CamVid\smash{$^{\dagger}$} & 162 & 265 & 325 & \textbf{0.92} & \textbf{0.90} & \textbf{0.89} & 0.88 & 0.90 & 0.89 \\
\midrule
\multicolumn{10}{@{}l}{\emph{Detection}} \\
VOC 2007\smash{$^{\dagger}$} & 391 & 771 & 1034 & \textbf{0.89} & \textbf{0.92} & \textbf{0.89} & 0.65 & 0.69 & 0.67 \\
DIOR\smash{$^{\dagger}$} & 371 & 558 & 638 & \textbf{0.78} & \textbf{0.76} & \textbf{0.78} & 0.60 & 0.71 & 0.68 \\
BUSI\smash{$^{\dagger}$} & 100 & 115 & -- & \textbf{0.82} & \textbf{0.78} & -- & 0.76 & 0.76 & -- \\
COCO\smash{$^{\dagger}$} & 358 & 628 & 784 & \textbf{0.86} & \textbf{0.85} & \textbf{0.86} & 0.63 & 0.65 & 0.71 \\
ExDark\smash{$^{\dagger}$} & 360 & 713 & 1005 & \textbf{0.92} & \textbf{0.92} & \textbf{0.94} & 0.70 & 0.71 & 0.70 \\
PlantDoc\smash{$^{\dagger}$} & 270 & 332 & -- & \textbf{0.94} & \textbf{0.92} & -- & 0.81 & 0.64 & -- \\
DocLayNet\smash{$^{\dagger}$} & 364 & 616 & 734 & \textbf{0.86} & \textbf{0.80} & \textbf{0.86} & 0.63 & 0.70 & 0.79 \\
\bottomrule
\end{tabular}
\end{table}